\documentclass[]{article} 
\usepackage{amsmath,amssymb}

\usepackage[table]{xcolor} 
\usepackage{tabularx}
\usepackage{multirow}
\usepackage{colortbl}

\usepackage{graphicx}
\usepackage{float}
\usepackage{subcaption} 
\usepackage[export]{adjustbox}
\usepackage[justification=centering]{caption} 
\usepackage{listings}
\usepackage{tcolorbox}
\tcbuselibrary{listings,breakable}

\usepackage{makeidx}
\usepackage{multicol}
\usepackage{lscape,rotating}
\usepackage{chngcntr}
\counterwithout{figure}{section}
\usepackage{capt-of} 

\usepackage{hyperref}
\usepackage[ruled,vlined,linesnumbered]{algorithm2e}
\SetKwComment{Comment}{$\triangleright$\ }{} 
\SetKwInput{KwIn}{Input}
\SetKwInput{KwOut}{Output}
\usepackage{placeins}
\usepackage{siunitx}
\usepackage{booktabs}

\usepackage{textcomp} 
\usepackage[utf8]{inputenc}
\usepackage[T1]{fontenc}
\usepackage{authblk}
\usepackage{textcomp}

\usepackage{lmodern}
\usepackage{tikz}
\usetikzlibrary{arrows.meta, positioning, shapes, calc, fit}

\newif\ifcolorfig
\colorfigtrue 

\ifcolorfig
  \definecolor{cBlue}{RGB}{33,150,243}
  \definecolor{cGreen}{RGB}{76,175,80}
  \definecolor{cOrange}{RGB}{255,152,0}
  \definecolor{cPurple}{RGB}{156,39,176}
  \definecolor{cIndigo}{RGB}{63,81,181}
  \newcommand{\boxfillA}{cBlue!8}\newcommand{\boxdrawA}{cBlue!70!black}
  \newcommand{\boxfillB}{cGreen!8}\newcommand{\boxdrawB}{cGreen!70!black}
  \newcommand{\boxfillC}{cOrange!10}\newcommand{\boxdrawC}{cOrange!80!black}
  \newcommand{\boxfillD}{cPurple!10}\newcommand{\boxdrawD}{cPurple!80!black}
  \newcommand{\boxfillE}{cIndigo!10}\newcommand{\boxdrawE}{cIndigo!80!black}
  \newcommand{\edgecol}{black!70}
\else
  \newcommand{\boxfillA}{gray!6}\newcommand{\boxdrawA}{black}
  \newcommand{\boxfillB}{gray!6}\newcommand{\boxdrawB}{black}
  \newcommand{\boxfillC}{gray!6}\newcommand{\boxdrawC}{black}
  \newcommand{\boxfillD}{gray!6}\newcommand{\boxdrawD}{black}
  \newcommand{\boxfillE}{gray!6}\newcommand{\boxdrawE}{black}
  \newcommand{\edgecol}{black}
\fi

\tikzset{
  stepbox/.style={
    rounded corners, draw, thick,
    minimum width=5.9cm, minimum height=3.0cm, 
    inner sep=3.2mm
  },
  paleblue/.style={fill=\boxfillA, draw=\boxdrawA},
  palegreen/.style={fill=\boxfillB, draw=\boxdrawB},
  paleyellow/.style={fill=\boxfillC, draw=\boxdrawC},
  palepurple/.style={fill=\boxfillD, draw=\boxdrawD},
  palelilac/.style={fill=\boxfillE, draw=\boxdrawE},
  arr/.style={-Latex, very thick, draw=\edgecol},
  dashedarr/.style={-Latex, thick, dashed, draw=\edgecol}
}

\tikzset{
  pics/phone/.style={
    code={
      \begin{scope}[line width=0.7pt, draw=\boxdrawA]
        \draw[rounded corners=0.8pt, fill=white] (0,0) rectangle (0.50,0.80);
        \fill (0.25,0.06) circle (0.018);
      \end{scope}
    }
  },
  pics/barchart/.style={
    code={
      \begin{scope}[line width=0.7pt, draw=\boxdrawB]
        \draw (0,0) rectangle (0.56,0.42);
        \fill[\boxdrawB] (0.08,0.08) rectangle (0.14,0.26);
        \fill[\boxdrawB] (0.24,0.08) rectangle (0.30,0.36);
        \fill[\boxdrawB] (0.40,0.08) rectangle (0.46,0.20);
      \end{scope}
    }
  },
  pics/brain/.style={
    code={
      \begin{scope}[line width=0.7pt, draw=\boxdrawC]
        \draw[rounded corners=2pt, fill=white] (0,0.10) .. controls (0,0.48) and (0.58,0.48) .. (0.58,0.10)
          .. controls (0.58,-0.12) and (0.0,-0.12) .. (0,0.10) -- cycle;
        \draw (0.14,-0.02) to[out=90,in=90] (0.44,-0.02);
        \draw (0.14,0.14) to[out=90,in=90] (0.44,0.14);
        \draw (0.29,-0.08) -- (0.29,0.24);
      \end{scope}
    }
  },
  pics/json/.style={
    code={
      \begin{scope}[line width=0.7pt, draw=\boxdrawD]
        \draw (0,0) -- (0.19,0) .. controls (0.31,0) and (0.31,0.30) .. (0.19,0.30) -- (0,0.30);
        \draw (0.60,0) -- (0.41,0) .. controls (0.29,0) and (0.29,0.30) .. (0.41,0.30) -- (0.60,0.30);
        \node[scale=0.68] at (0.30,0.15) {\{\}};
      \end{scope}
    }
  },
  pics/graph/.style={
    code={
      \begin{scope}[line width=0.7pt, draw=\boxdrawE]
        \fill (0.06,0.06) circle (0.022);
        \fill (0.52,0.06) circle (0.022);
        \fill (0.18,0.36) circle (0.022);
        \fill (0.40,0.32) circle (0.022);
        \draw (0.06,0.06) -- (0.18,0.36) -- (0.40,0.32) -- (0.52,0.06) -- (0.06,0.06);
        \draw (0.18,0.36) -- (0.52,0.06);
      \end{scope}
    }
  }
}

\usepackage{fancyhdr}
\let\origtitle\title 
\renewcommand{\title}[1]{\lfoot{\textit{#1}}\origtitle{\textbf{#1}}}
\renewcommand{\sectionmark}[1]{\markboth {}{}}

\date{}

\providecommand{\keywords}[1]
{
  \small	
  \textbf{\textit{Keywords---}} #1
}

\title{LLM-Guided Exemplar Selection for Few-Shot Wearable-Sensor Human Activity Recognition}

\author[1,2]{Elsen Ronando\footnote{\href{mailto:ronando.elsen840@mail.kyutech.jp}{ronando.elsen840@mail.kyutech.jp}}}
\author[1]{Sozo Inoue \footnote{\href{mailto:sozo@brain.kyutech.ac.jp}{sozo@brain.kyutech.ac.jp}}}
\affil[1]{Graduate School of Life Science and Systems Engineering, Kyushu Institute of Technology, Kitakyushu, Japan}
\affil[2]{Department of Informatics, Universitas 17 Agustus 1945 Surabaya, Surabaya, Indonesia}

\begin{document}
\maketitle
\thispagestyle{fancy}
\centering

\abstract{
In this paper, we propose an LLM-Guided Exemplar Selection framework to address a key limitation in state-of-the-art Human Activity Recognition (HAR) methods: their reliance on large labeled datasets and purely geometric exemplar selection, which often fail to distinguish similar weara-ble sensor activities such as walking, walking upstairs, and walking downstairs. Our method incorporates semantic reasoning via an LLM-generated knowledge prior that captures feature importance, inter-class confusability, and exemplar budget multipliers, and uses it to guide exemplar scoring and selection. These priors are combined with margin-based validation cues, PageRank centrality, hubness penalization, and facility-location optimization to obtain a compact and informative set of exemplars. Evaluated on the UCI-HAR dataset under strict few-shot conditions, the framework achieves a macro F1-score of 88.78\%, outperforming classical approaches such as random sampling, herding, and $k$-center. The results show that LLM-derived semantic priors, when integrated with structural and geometric cues, provide a stronger foundation for selecting representative sensor exemplars in few-shot wearable-sensor HAR.
}


\section{Introduction}
\label{section:Introduction}

Human Activity Recognition (HAR) has become a central topic in pervasive and wearable computing, supporting applications in healthcare, rehabilitation, elderly monitoring, and human–computer interaction \cite{LI202347, 9257355, bios12060393, YAZICI2023100971, Eldrandaly2023ActivBench:}. Although deep-learning models achieve strong performance on large annotated datasets \cite{balaha_comprehensive_2023, s23052816}, real-world deployment often faces limited labeled data due to the cost of collecting long sensor recordings and producing reliable annotations \cite{s24155045, iot1020025}. Domain variability across users, devices, and environments further challenges model generalization \cite{9913339, 9508415}.

\begin{figure}[H]
\centering
\begin{subfigure}[t]{.5\linewidth}
\includegraphics[width=\linewidth]{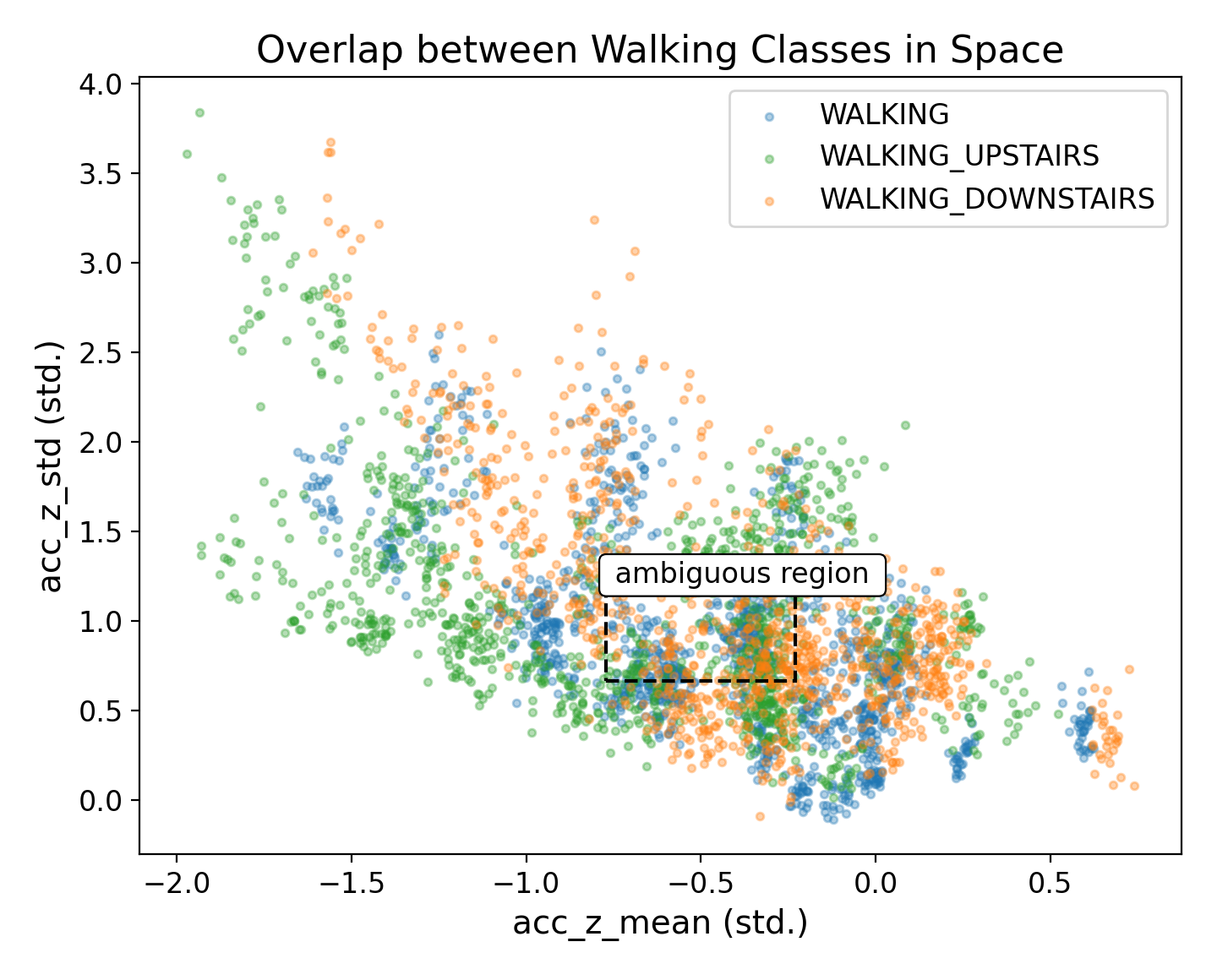}
\caption{Ambiguity among walking activities}
\end{subfigure}\hfill
\begin{subfigure}[t]{.5\linewidth}
\includegraphics[width=\linewidth]{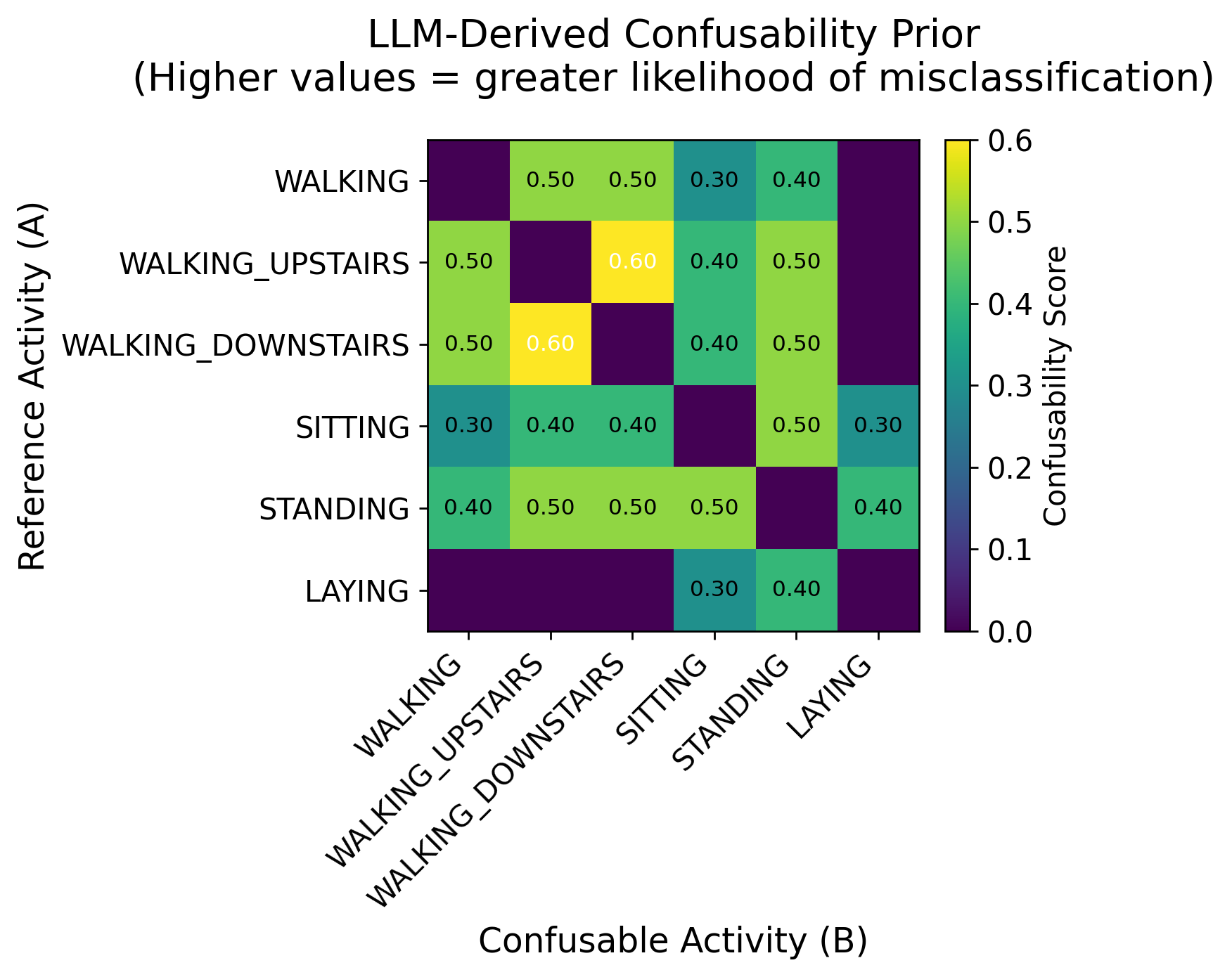}
\caption{LLM confusability prior}
\end{subfigure}
\caption{Motivation for the proposed LLM-Guided Exemplar Selection.
(a) Overlapping signal distributions cause Ambiguity between similar walking activities.
(b) LLM-derived semantic priors reflect conceptual closeness (e.g., \textit{walking} vs.\ \textit{walking upstairs}).}
\label{fig:intro_motivation}
\end{figure}

A common strategy to reduce data requirements is \emph{exemplar selection}, in which a compact set of representative samples is selected for training. Classical approaches such as random sampling, herding, and $k$-center rely purely on geometric distances \cite{BELOUADAH202138, ramalingam2023weightedkcenteralgorithmdata}. However, they ignore semantic relationships among activities, leading to confusion when their signal patterns overlap, for instance, among walking-related activities. Fig.~\ref{fig:intro_motivation} illustrates this challenge: overlapping distributions (a) and conceptual closeness captured by LLM-based confusability (b) both highlight the need for selection methods that incorporate semantic reasoning.

Large Language Models (LLMs) offer new opportunities to incorporate semantic reasoning into HAR. Recent works show that LLMs can generate structured knowledge and support multimodal inference \cite{ElsenRonando202418, dagdelen_structured_2024, 10.1145/3663741.3664785}, but their integration into sensor-based few-shot learning remains limited. Early attempts, such as HED-LM \cite{s25113324}, used LLM scoring to complement geometric distances but treated samples independently and lacked mechanisms to encode global structure, inter-class relations, or semantic priors. These gaps motivate a more principled integration of reasoning-guided signals into exemplar selection.

To address these limitations, this study proposes an \textit{LLM-Guided Exemplar Selection} framework that extends LLMs' role from local scoring to generating semantic features, structured domain knowledge, and reasoning cues, refined using PageRank-based centrality and hubness adjustment \cite{HASHEMI2020113024, math13071202}. This formulates exemplar selection as a hybrid semantic–structural process rather than a purely geometric one.

The following research questions guide the study:
\begin{enumerate}
\item \textbf{RQ1:} How can LLMs generate meaningful semantic features from numerical sensor data for few-shot HAR?
\item \textbf{RQ2:} How can structured domain knowledge produced by LLMs be incorporated into exemplar evaluation?
\item \textbf{RQ3:} How can semantic reasoning and structural regularization be combined to select diverse and representative exemplars?
\end{enumerate}

The key contributions of this study can be outlined as follows:
\begin{enumerate}
\item A method for deriving LLM-based semantic features that capture activity-relevant dimensions beyond statistical descriptors.
\item A mechanism that integrates structured LLM knowledge, feature weights, inter-class confusion, and exemplar budgets, into a unified scoring formulation.
\item A hybrid exemplar selection algorithm combining validation margin, PageRank centrality, hubness penalties, and LLM priors within a facility-location framework.
\end{enumerate}

We evaluate the framework on the UCI-HAR dataset \cite{Anguita2013APD} under strict few-shot settings. The suggested approach reaches a macro F1-score of 88.78\%, consistently outperforming random, herding, and $k$-center baselines across multiple classifiers. Ablation studies further confirm the complementary roles of semantic features, structured LLM knowledge, and hybrid selection.

\textbf{The remainder of this paper is organized as follows.}
Section~\ref{section: Related} reviews related work on few-shot HAR, exemplar selection, and LLM integration. Section~\ref{section:proposed} presents the proposed framework. Section~\ref{section:experiment} describes the experimental setup and results. Section~\ref{section:discussion} provides further discussion, and Section~\ref{section:conclusion} concludes the paper.
\section{Related Work}
\label{section: Related}
Human Activity Recognition (HAR) has gained substantial attention over the past decade, in parallel with the widespread use of wearable and mobile sensing devices. Using motion data from sensors such as accelerometers and gyroscopes, these systems attempt to identify and classify daily human activities under realistic conditions \cite{LI202347, 9257355}. Applications of HAR now span a broad range of areas, from healthcare and physical rehabilitation to elderly care and innovative environments \cite{bios12060393, YAZICI2023100971, Eldrandaly2023ActivBench:}. Although the field has seen impressive growth, the reliability of existing models still depends on the availability of large, annotated datasets. In practice, collecting such datasets is not trivial. It requires long recording sessions, manual labeling, and substantial human effort. Moreover, differences in user behavior, device placement, and environmental context often make the data highly variable, which poses a serious challenge to model generalization. These limitations have prompted researchers to seek methods that work effectively even when labeled data are scarce. This is where few-shot learning (FSL) has begun to play an increasingly important role.

Few-shot learning enables a model to recognize unseen or underrepresented activity classes with only a few labeled samples. Within the HAR domain, several strategies have been explored, including metric-based prototypical-style methods, transfer learning, and various data augmentation schemes designed to simulate variability \cite{9767306, 10.1145/3704921, 9712302}. These approaches have improved recognition performance under limited supervision, yet their effectiveness still hinges on the quality of the available examples. In few-shot scenarios, every single training sample carries significant influence, meaning that one unrepresentative or noisy instance can distort the learned decision boundaries \cite{s25113324, pecher2024automaticcombinationsampleselection}. This makes the selection of representative and diverse exemplars a critical step in achieving stable performance.

Traditional exemplar selection techniques (e.g., random sampling, herding, k-center) and more advanced instance selection pipelines \cite{PANJA2023119536} mainly rely on geometric characteristics of the feature space. While these methods are computationally simple and widely used, they often ignore the semantic meaning of activities. For example, signals generated by walking and walking upstairs may appear close in geometric space, even though the two activities are conceptually distinct. As a result, models trained on such exemplars may misclassify activities that share similar motion patterns. To mitigate this, recent works have introduced additional structural information, such as graph connectivity \cite{Dong_Hong_Tao_Chang_Wei_Gong_2021}, graph-based weighting \cite{ZHOU2023230}, and inter-class margin analysis \cite{RAN2024103664}. These methods allow a model to account for broader relationships between samples rather than relying solely on pairwise distances. However, they still rely heavily on geometric or statistical cues and rarely engage in semantic reasoning, limiting their ability to interpret data at a conceptual level. This gap provides an opportunity for integrating reasoning-driven models, such as Large Language Models (LLMs).

The emergence of LLMs has introduced new possibilities for reasoning with heterogeneous and multimodal data. Beyond their success in text processing, LLMs have demonstrated an ability to link numeric and symbolic representations, allowing them to describe how sensor signals relate to human behavior in abstract terms \cite{ElsenRonando202418, li2025sensorllmaligninglargelanguage,oai:ipsj.ixsq.nii.ac.jp:00239430}. For example, an LLM can infer that variations in vertical acceleration may correspond to transitions between walking and running. Recent studies have begun to explore this potential by using LLMs for feature reasoning or to produce structured domain knowledge \cite{shorten2024structuredragjsonresponseformatting,s25113324} in the form of feature weights, inter-class similarity matrices, and confusion relationships. Nevertheless, most of these efforts still treat the language model as a supporting tool, providing textual judgments or scoring hints, rather than as an integrated component within a sensor-learning pipeline. This observation motivates our work to go a step further: to design an LLM-guided exemplar selection framework that blends semantic reasoning with geometric regularization for few-shot HAR.

In our approach, the LLM plays an active role instead of remaining a passive evaluator. It contributes to shaping the feature space, assigning class-specific priors, and influencing which samples are selected as exemplars. The semantic priors obtained from the LLM are then refined using structural regularization techniques, such as PageRank and hubness adjustment, thereby balancing conceptual relevance and geometric consistency. By combining these two aspects, the framework links symbolic reasoning with learning numerical representations.

To summarize, most earlier studies on few-shot HAR have been grounded in either statistical or geometric reasoning. In contrast, recent attempts to introduce LLMs have not yet fully leveraged their reasoning capabilities. Our study seeks to bridge this gap by unifying semantic understanding and structural regularization. Through this integration, we aim to enhance exemplar representativeness and improve generalization when only a small number of labeled samples are available.
\section{Proposed Method}
\label{section:proposed}
This section describes the proposed LLM-Guided Exemplar Selection Framework for few-shot human activity recognition (HAR). The main objective of this method is to improve activity recognition performance when only a small amount of labeled data is available. The framework combines numerical features extracted from sensor signals with semantic reasoning provided by a Large Language Model (LLM). It operates through three main stages: exemplar selection, model training, and Inference, each designed to enhance the system's representativeness and generalization.

\subsection{Overview of the Framework}
\label{over_framework}
\begin{figure}[ht!]
\centering
\includegraphics[width=\linewidth]{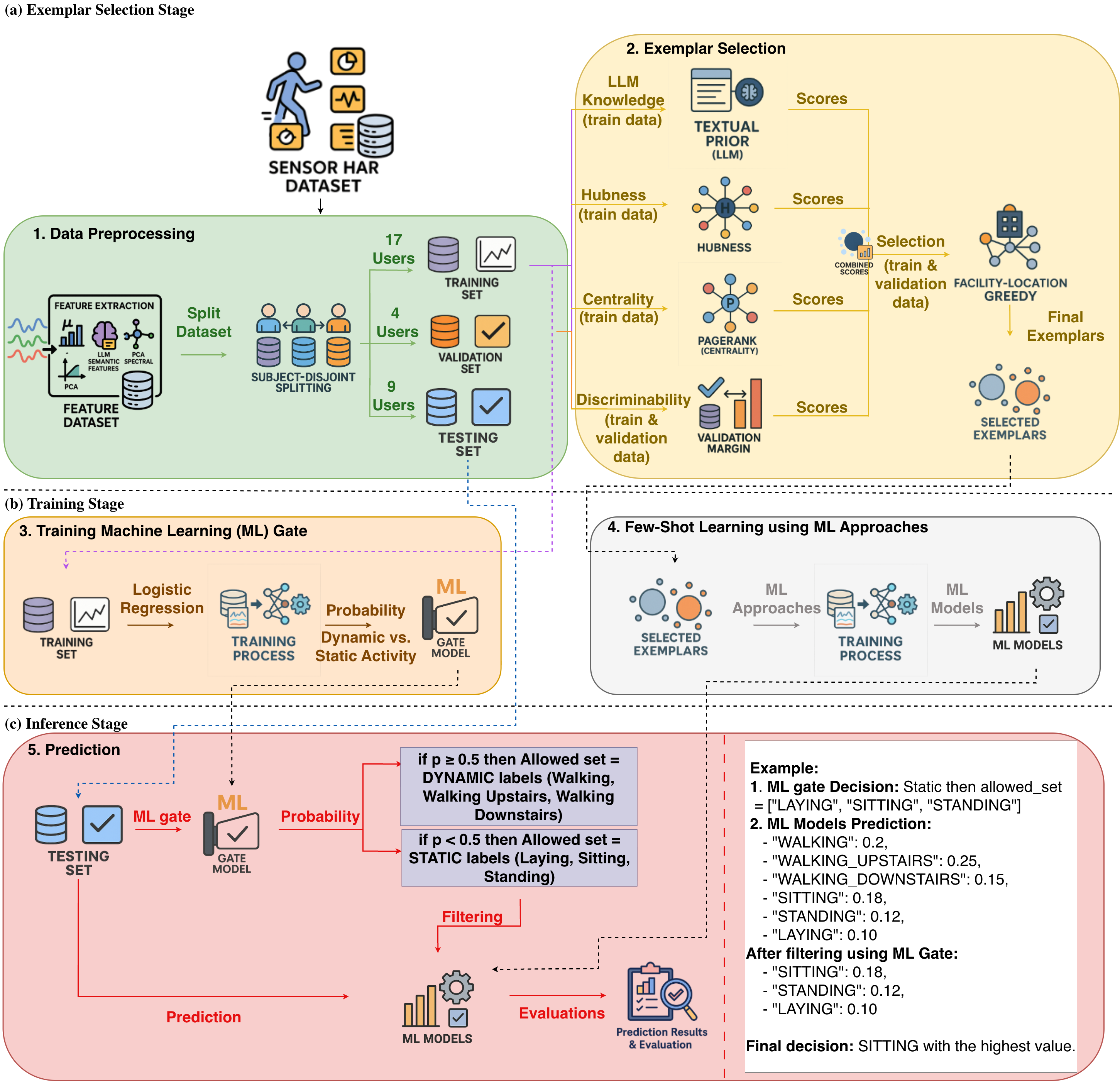}
\caption{Overall workflow of the proposed LLM-Guided Exemplar Selection Framework. 
The process consists of three main stages:  (a) Exemplar Selection, which combines LLM-guided semantic reasoning and facility-location selection;  (b) Training, where selected exemplars are used to train the ML models; and (c) Inference: the ML Gate sends each test sample to the static or dynamic ML 
models, and the arrows show which model group produces the final prediction.}
\label{fig:pipeline-llm-har}
\end{figure}
Figure~\ref{fig:pipeline-llm-har} outlines our method in three stages. 
(a) \textit{Exemplar Selection}: we preprocess the UCI-HAR signals, apply a subject-disjoint split, and compute basic (time/frequency) features. An LLM then supplies two inputs, semantic feature axes and a compact knowledge object (label--feature weights, inter-class confusability, and class budget multipliers). Each training sample is scored by a hybrid criterion (validation margin, class-conditional PageRank, hubness penalty, and the LLM prior), and the final per-class exemplars are chosen via greedy facility location. 
(b) \textit{Training}: using only the selected exemplars, we train downstream ML models, and then using the training data, we train a lightweight gate that routes test windows into dynamic vs.\ static activity sets.
(c) \textit{Inference}: the gate filters the candidate label set for each test window; the ML models then predict within the allowed set, and we report the final decision. This design keeps computation small, prevents leakage across splits, and injects semantic reasoning where it matters most, at exemplar selection.

\subsection{Exemplar Selection Stage}
\label{exemplar_selection}
The exemplar selection stage is the core component of the proposed framework. This stage aims to identify a small but representative subset of training samples that can effectively describe each activity class. Unlike traditional distance-based methods, the proposed approach integrates semantic information from a Large Language Model (LLM) and structural relationships among samples to improve representativeness and diversity. The process consists of five main steps: data preprocessing, feature extraction, LLM knowledge encoding, hybrid scoring, and facility-location-based selection.

\noindent\textbf{Data Preprocessing}. The preprocessing stage aims to convert raw inertial signals into structured features that the proposed framework can directly use. As illustrated in Fig.~\ref{fig:pipeline-llm-har}, this stage begins with feature extraction followed by dataset partitioning. Conceptually, all recorded signals are first transformed into informative numerical representations, such as statistical, spectral, and semantic features, that capture the underlying characteristics of each activity. These features are then organized into training, validation, and testing subsets for subsequent exemplar selection and model evaluation. The subject-wise division used in this study is summarized in Table~\ref{tab:data_split_simple}.
\begin{table}[h!]
\centering
\caption{Subject-wise data division used in the experiments.}
\resizebox{\linewidth}{!}{
\begin{tabularx}{\linewidth}{lccc}
\hline
\textbf{Subset} & \textbf{Samples} & \textbf{Subjects} & \textbf{Subject IDs} \\
\hline
Training & 6{,}012 & 17 & [1, 3, 6, 8, 11, 14--17, 19, 21, 23, 26--30] \\
Validation & 1{,}340 & 4 & [5, 7, 22, 25] \\
Testing & 2{,}947 & 9 & [2, 4, 9--10, 12--13, 18, 20, 24] \\
\hline
\end{tabularx}
}
\label{tab:data_split_simple}
\end{table}

The study employs the UCI Human Activity Recognition (HAR) dataset, which includes tri-axial accelerometer and gyroscope data collected from 30 volunteers performing six common activities: walking, walking upstairs, walking downstairs, sitting, standing, and laying \cite{Anguita2013APD}. Each continuous signal is divided into fixed-length windows of 2.56 seconds (128 samples) with 50\% overlap, yielding multiple segments per activity. This segmentation approach preserves temporal continuity and enhances the statistical reliability of extracted features.

For each window, time-domain features are derived, including the mean and standard deviation of six sensor channels: (acc\_x, acc\_y, acc\_z, gyr\_x, gyr\_y, gyr\_z). These basic descriptors serve as the foundation for subsequent transformations, including Principal Component Analysis (PCA), Spectral PCA, and LLM-based semantic feature generation. All features are normalized using z-score standardization to ensure consistency across users and sensor dimensions.

\noindent\textbf{Feature Construction and Representation}. After preprocessing, the next stage focuses on constructing representative features that capture both the physical characteristics of human motion and high-level semantics. Four complementary feature sets are generated: (1) statistical features in the time domain, (2) Principal Component Analysis (PCA) features, (3) Spectral PCA features in the frequency domain, and (4) semantic features derived from a Large Language Model (LLM). These feature groups are concatenated into a unified representation, which serves as the input space for exemplar selection and classification.

\begin{enumerate}
    \item \textbf{Statistical features in the time domain}. For each 2.56 s window (128 samples), the mean and standard deviation are computed for six inertial channels: (acc\_x, acc\_y, acc\_z, gyr\_x, gyr\_y, gyr\_z). These parameters describe average movement intensity and signal variability, providing a compact summary of temporal dynamics.
    \item \textbf{Principal Component Analysis (PCA)}. PCA is applied to reduce feature dimensionality while preserving major variance components. Six principal components are retained, capturing the most salient correlations among sensor axes and minimizing redundancy.
    \item \textbf{Spectral PCA}. To complement time-domain statistics, each signal window is transformed into the frequency domain through a real Fast Fourier Transform (rFFT). The resulting log-amplitude spectra from the six channels are merged and projected using PCA until 95\% of the total variance is preserved. These components represent dominant rhythmic patterns associated with cyclic movements such as walking or walking upstairs.
    \item \textbf{LLM-based semantic features}. Beyond numerical transformations, an LLM is employed to generate additional semantic dimensions. The model analyzes per-class feature statistics and produces compact linear combinations of base channels that encode activity-related semantics (e.g., verticality, balance, motion strength). Each semantic feature is expressed as a weighted sum of standardized sensor attributes, forming interpretable axes that link sensor behavior to conceptual movement patterns. The LLM prompting design is shown in Fig. \ref{fig:LLM-Semantic-Feature}. More details of the algorithm are shown in Algorithm \ref{algo_b1}. For instance, one example of the generated semantic features is the verticality axis, which the LLM defines as a weighted combination of z-axis acceleration statistics, e.g., verticality = 0.8 $\times$ acc\_z\_mean $+$ 0.6 $\times$ acc\_z\_std, representing the intensity of upward–downward body motion. Additional examples and the full JSON specification are provided in Appendix~\ref{semantic}.
\end{enumerate}

\noindent\textbf{LLM Knowledge Encoding}. In this stage, a Large Language Model (LLM) is employed to generate structured domain knowledge that serves as a semantic prior for exemplar evaluation, with the prompting design shown in Fig. \ref{fig:LLM-Know-Prompting}. Unlike traditional approaches that rely solely on geometric or statistical information, the proposed framework integrates conceptual understanding by encoding relationships between activities and their corresponding sensor features.

The LLM receives summarized information from the training subset, including class-wise mean values of standardized features. Through carefully designed prompts, it produces a compact JSON-based knowledge representation consisting of three key components:
\begin{itemize}
    \item Feature weights per class (label\_feature\_weights) represent the relative importance of each feature in identifying a specific activity. For example, higher weights on acc\_z\_mean and gyr\_y\_std are typically associated with WALKING\_UPSTAIRS, which involves strong vertical acceleration and angular rotation.
    \item Inter-class confusability matrix (confusability) captures semantic similarity or potential overlap between activity classes. For instance, WALKING and WALKING\_DOWNSTAIRS often exhibit a confusability value around 0.8, indicating high contextual similarity in sensor patterns.
    \item Label budget multiplier (label\_budget\_multiplier) defines how many exemplars should be selected for each class, adjusted according to the activity’s complexity. Dynamic activities (e.g., WALKING and WALKING\_UPSTAIRS) receive higher multipliers ($\simeq 1.2\times$), while static ones (e.g., LAYING, SITTING) are assigned lower multipliers ($\simeq 0.8\times$).
\end{itemize}

We include LLM-based knowledge encoding because geometric features alone cannot capture the actual meaning of each activity. Some activities, such as walking and walking downstairs, often appear similar in the feature space, even though they differ in practice. The knowledge provided by the LLM helps the scoring process understand which features truly matter for each activity and which activity pairs are naturally more confusable. This additional semantic information guides the exemplar selection process and reduces the risk of selecting samples that look correct numerically but are not representative in a real-world sense.

All three components are stored as a single structured object, referred to as LLM Knowledge JSON, and are directly integrated into the exemplar scoring process. This semantic prior enables the selection algorithm to interpret candidate exemplars not only by spatial proximity but also by their conceptual consistency within the activity domain. Formally, for each training sample $i$ with feature vector $\mathbf{x}_i$ and class label $y_i$, the LLM-based semantic score $T(i)$ is computed in Eq. \ref{eqn:ti}.
\begin{equation}\label{eqn:ti}
    T(i) = \mathbf{w}_{y_i}^\top \mathbf{x}i - \max_{y' \neq y_i} c_{y_i,y'} \, (\mathbf{w}_{y'}^\top \mathbf{x}_i)
\end{equation}
where $\mathbf{w}_{y_i}$ denotes the feature-weight vector for class $y_i$, and $c_{y_i,y’}$ represents the confusability between class $y_i$ and $y’$. A higher $T(i)$ value indicates that the exemplar is semantically representative of its class while remaining distinct from other conceptually similar activities. By introducing LLM-guided knowledge encoding as illustrated in Algorithm \ref{algo_b2}, the framework effectively bridges numerical features and semantic reasoning, forming the foundation for the Hybrid Scoring Mechanism. The structured LLM response containing the label-feature weights, class confusability, and label budget multipliers used in this stage is presented in Appendix~\ref{prior}.

\noindent\textbf{Hybrid Scoring Mechanism}. This stage defines the hybrid scoring function used to evaluate each candidate exemplar within the training subset. The objective is to quantify how representative, distinctive, and semantically relevant a sample is, combining both data-driven and reasoning-based cues. The overall score integrates four complementary factors: (1) validation margin \cite{liu2020negativemarginmattersunderstanding}, (2) PageRank-based centrality \cite{HASHEMI2020113024}, (3) hubness penalty \cite{math13071202}, and (4) semantic prior from the LLM. For each candidate $i$ with feature vector $\mathbf{x}_i$ and label $y_i$, the total hybrid score is formulated in Eq. \ref{eqn:feature_vector}.
\begin{equation}\label{eqn:feature_vector}
    S(i) = \alpha M(i) + \mu P(i) - \tau H(i) + \beta T(i)
\end{equation}
where:
\begin{itemize}
    \item $M(i)$: positive–negative margin score based on candidate similarity to validation data (same class vs. different class),
    \item $P(i)$: graph-based PageRank score that assesses the centrality of exemplars in the feature space,
    \item $H(i)$: hubness penalty to avoid candidates that are too often neighbors of many other samples,
    \item $T(i)$: semantic score from LLM knowledge defined in Eq. \ref{eqn:ti}.
\end{itemize}
The coefficients $\alpha$, $\mu$, $\tau$, $\beta$ balance the contribution of each component and are determined empirically through validation ($\alpha\!=\!1.0,\;\mu\!=\!0.10,\;\tau\!=\!0.10,\;\beta\!=\!0.15$ in this study).

In practice, the validation margin M(i) is computed using cosine similarity between the candidate and validation subsets of positive and negative classes in Eq. \ref{eqn:mi}. 
\begin{equation}\label{eqn:mi}
    M(i)=\frac{1}{|V^+|}\sum_{j\in V^+}\text{sim}(\mathbf{x}_i,\mathbf{x}j)
-\frac{1}{|V^-|}\sum{k\in V^-}\text{sim}(\mathbf{x}_i,\mathbf{x}_k)
\end{equation}
where $V^+$ and $V^-$ denote validation sets belonging to the same and different classes, respectively. PageRank values $P(i)$ are estimated from a class-conditional mutual-kNN graph using cosine edge weights. In contrast, hubness scores $H(i)$ are obtained by counting the number of times each sample appears in the neighbor sets of others. 

This hybrid formulation ensures that exemplar evaluation reflects both geometric structure (through margin and centrality) and semantic reasoning (through LLM priors), providing a balanced trade-off between representativeness and conceptual discriminability. The resulting scores serve as input to the exemplar selection algorithm described in Algorithm \ref{algo_b3}.

\noindent\textbf{Facility-Location-Based Selection}. After computing the hybrid scores for all candidate samples, the final step selects a subset of exemplars that best represent each activity class. To achieve both coverage and diversity, the selection process is formulated as a submodular optimization problem using the Facility Location criterion \cite{kaushal2019learningdataunifieddata}. Given the set of training samples $X = \{x_1, x_2, \dots, x_n\}$, the goal is to select a subset $S \subseteq X$ that maximizes the overall representativeness function in Eq. \ref{eqn:facility}. The algorithm for this stage is described in detail in Algorithm \ref{algo_b4}.
\begin{equation}\label{eqn:facility}
    F(S)=\sum_{x\in X}\max_{s\in S}\text{sim}(x,s),
\end{equation}
where $\text{sim}(x, s)$ denotes the cosine similarity between sample $x$ and selected exemplar $s$. This formulation ensures that each data point in the class is well represented by at least one exemplar in $S$. To efficiently solve this optimization, a greedy algorithm is applied. At each iteration, the candidate $s^*$ that provides the highest incremental gain in $F(S)$ is selected in Eq. \ref{eqn:greedy}
\begin{equation}\label{eqn:greedy}
    s^* = \arg\max_{s \in X \setminus S} \big[ F(S \cup \{s\}) - F(S) \big].
\end{equation}
The algorithm continues until the exemplar budget per class is reached. The budget is determined adaptively based on the label budget multiplier generated by the LLM Knowledge Encoding. Specifically, dynamic activities such as WALKING and WALKING\_UPSTAIRS receive a larger number of exemplars (e.g., $k_{\text{dyn}} = 8$). In contrast, static activities such as SITTING and LAYING use smaller budgets (e.g., $k_{\text{stat}} = 2$). Through this integration, the Facility-Location stage transforms exemplar selection into a global optimization problem, in which geometric structure and semantic priors jointly determine the most informative and diverse exemplars for each activity class.

\subsection{Training Stage}
\label{training}
The training stage begins after exemplar selection is completed. Before model training, a Machine Learning (ML) Gate is applied to categorize activities into two main groups: static (e.g., sitting, standing, laying) and dynamic (e.g., walking, walking upstairs, walking downstairs). This gating mechanism ensures that each classifier is trained only on data relevant to its activity type, thereby reducing inter-class confusion and improving model stability.

For each gate, the selected exemplars are used to train several baseline models, including k-Nearest Neighbors (k-NN), Logistic Regression, Random Forest, Histogram Gradient Boosting, Linear SVC, and Gaussian Naïve Bayes. All models are trained on standardized feature representations that combine both statistical and semantic features derived from the LLM-guided pipeline.
The ML Gate outputs are later merged during the inference stage to produce unified predictions across all activity classes. No information from validation or test subsets is used at this stage to prevent data leakage.

\subsection{Inference Stage}
\label{inference}
During inference, each test instance undergoes the same preprocessing, feature extraction, and semantic transformation steps as in training. The ML Gate first determines whether the incoming signal belongs to a static or dynamic activity domain. This step directs the sample to the corresponding trained model group. Once gated, the prediction is made based on similarity to the selected exemplars or the decision boundaries of the trained classifier within that group. The final predicted label is generated by combining the outputs of both gates to produce a complete six-class HAR result. This design allows the inference process to be both efficient and robust, since the ML Gate minimizes confusion between distinct motion types and ensures that each classifier focuses on a specific behavior domain, as is shown in Appendix \ref{mlgate}.

In summary, the proposed LLM-Guided Exemplar Selection framework enhances few-shot HAR by integrating semantic reasoning, hybrid scoring, and structured model training. The process begins with preprocessing and feature formation, followed by exemplar selection guided by LLM knowledge. A Machine Learning Gate then divides activities into static and dynamic domains, ensuring more focused model training. Finally, a facility-location strategy ensures that each exemplar subset is both representative and diverse. This integration enables efficient recognition with minimal labeled data while maintaining high interpretability and generalization to unseen users. ore details of the algorithm are illustrated in Algorithm \ref{algo_b5}.
\section{Experimental Design and Evaluation}
\label{section:experiment}
This section presents the experimental setup, evaluation procedures, and main findings of applying the proposed LLM-Guided Exemplar Selection framework on the UCI-HAR dataset. The objective is to verify its effectiveness in few-shot Human Activity Recognition (HAR) compared to classical baselines.

\subsection{Experimental Objectives and Design}
\label{eo_des}
The experiment aims to evaluate how the proposed framework improves few-shot HAR through: (1) semantic feature generation from sensor data using LLMs; (2) integration of structured LLM knowledge into exemplar scoring; and (3) hybrid reasoning that fuses semantic, geometric, and graph-based cues for better generalization.

A subject-independent setting was adopted to ensure fair evaluation: training data were used for model construction, validation data for margin and graph computation, and testing data for unseen users. A Machine Learning Gate (ML Gate) was also implemented as a binary classifier separating static (LAYING, SITTING, STANDING) and dynamic (WALKING, WALKING\_UPSTAIRS, WALKING\_DOWNSTAIRS) activities. During inference, the gate restricts the label space, allowing only relevant activity subsets to be considered, thereby reducing confusion between motion types.

The experiment proceeded in two stages:
\begin{enumerate}
    \item LLM Knowledge Generation: Produces three key priors, (a) label-feature weights, (b) inter-class confusability, and (c) label budget multipliers, each encoded in JSON for numerical integration.
	\item Hybrid Exemplar Selection: Combines inter-class margins, PageRank, hubness penalties, and LLM priors within a facility-location framework to select representative, diverse exemplars.
\end{enumerate}

Each exemplar subset was then used to train six classical models: kNN, Logistic Regression, Random Forest, Gradient Boosting, Linear SVC, and Gaussian Naïve Bayes. Performance was evaluated using macro F1-score, comparing against Random, Herding, and K-Center baselines. An ablation study was also conducted to quantify the contributions of key components (LLM prior, graph structure, hubness, and facility location).

\subsection{Dataset and Experimental Settings}
\label{data_experiment}
The experiments used the UCI-HAR dataset \cite{Anguita2013APD}, which contains tri-axial accelerometer and gyroscope data from 30 participants performing six daily activities. Data splitting was subject-independent: 17 for training, 4 for validation, and 9 for testing.

All features were standardized (z-score) and combined into a multimodal representation including statistical, spectral (PCA), and semantic (LLM-based) features. The main parameters were fixed as: $k_{\text{graph}}=10,\;\lambda_{\text{red}}=0.25,\;\alpha_{\text{prior}}=0.20,\;\mu_{\text{cent}}=0.10,\;\tau_{\text{hub}}=0.10$. The entire pipeline was executed with seed = 42 for reproducibility. The ML Gate was trained on all combined features to distinguish static and dynamic domains before classifier inference.

In this study, we use the \texttt{GPT-4o-mini} model (temperature $0$, maximum tokens up to 1200) via LangChain to generate both the semantic feature axes and the knowledge JSON, using only training-set statistics in the prompts to ensure that no information from the validation or test sets leaks into the LLM.

\subsection{Experimental Results and Analysis}
\label{ex_an}
\subsubsection{Main Results}
\label{main_results}
Table~\ref{tab:model_comparison} summarizes macro F1-scores across six models. The proposed method consistently outperformed all baselines, achieving the highest mean score (67.02\%), a +10.36 point gain over Random Sampling.
\begin{table}[t]
\centering
\caption{Comparison of few-shot selection strategies across classifiers (Macro F1, \%). The \textit{Improvement} column is the gain of the proposed LLM-Guided approach relative to Random Sampling. More details of the confusion matrix of our approach are shown in the Appendix \ref{confusion} in Fig. \ref{fig:conf_matrix_ours_all}.}
\label{tab:model_comparison}

\renewcommand{\arraystretch}{1.2}
\setlength{\tabcolsep}{3pt}

\begin{tabularx}{\linewidth}{
>{\raggedright\arraybackslash}X
*{5}{>{\centering\arraybackslash}X}
}
\toprule
\textbf{Model} & \textbf{Random Sampling} & \textbf{Herding} \cite{BELOUADAH202138} & \textbf{K-Center} \cite{ramalingam2023weightedkcenteralgorithmdata} & \textbf{LLM-Guided (Ours)} & \textbf{Improvement} \\
\midrule
kNN           & 58.64 & \underline{74.97} & 60.72 & \textbf{85.88} & +27.24 \\
Logistic Regression    & 68.79 & 79.64 & \underline{81.12} & \textbf{88.79} & +19.99 \\
Random Forest          & 66.48 & 73.69 & \underline{75.55} & \textbf{79.61} & +13.13 \\
HistGradient Boosting  & 16.99 & 16.99 & 16.99 & 16.99 & +0.00 \\
Linear SVC             & \underline{74.45} & \textbf{78.50} & 71.23 & 72.85 & -1.60 \\
Gaussian Naïve Bayes & 54.64 & 45.15 & \textbf{70.70} & \underline{58.03} & +3.39 \\
\midrule
Mean          & 56.66 & 61.49 & \underline{62.72} & \textbf{67.02} & \textbf{+10.36} \\
\bottomrule
\end{tabularx}
\end{table}

The most significant improvements were observed in kNN (+27.24) and Logistic Regression (+19.99), demonstrating the synergy between semantic reasoning and geometric representation. The ML Gate, powered by Logistic Regression, also contributed to higher precision by minimizing inter-domain confusion. Detailed UMAP-based visualizations of exemplar distribution, including per-class and 
global projections, are presented in Appendix \ref{visualization} for further analysis. In addition to accuracy improvements, we also evaluated the computational efficiency of the proposed framework. As shown in Appendix~\ref{time}, the inference runtime remains comparable to the baseline exemplar-selection methods, confirming that the integration of LLM-guided exemplar selection does not introduce additional latency.

\subsubsection{Component Contribution (Ablation Study)}
\label{ablation}
To analyze component importance, each element of the hybrid scoring function was removed sequentially. Table~\ref{tab:ablation_study} shows that removing any component reduced accuracy, confirming their complementary roles.
\begin{table}[t]
\centering
\caption{Ablation study on key components of the LLM-Guided Exemplar Selection (Macro F1, \%).}
\label{tab:ablation_study}
\renewcommand{\arraystretch}{1.2}
\setlength{\tabcolsep}{4pt}

\begin{tabularx}{\linewidth}{
>{\raggedright\arraybackslash}X
>{\centering\arraybackslash}X
>{\centering\arraybackslash}X
}
\toprule
\textbf{Configuration} & \textbf{Mean Macro F1} & \textbf{Change} \\
\midrule
Full System (All Components) & \textbf{67.02} & — \\
Without Semantic Prior (LLM Knowledge) & 64.97 & –2.05 \\
Without Class-Conditional PageRank & 65.44 & –1.58 \\
Without Hubness Penalty & 65.70 & –1.32 \\
Without Facility Location Greedy & 47.34 & –19.68 \\
\bottomrule
\end{tabularx}
\end{table}

The Facility-Location step proved most critical (–19.68 points), followed by the LLM Prior (–2.05). This indicates that semantic knowledge improves contextual understanding, while facility-location ensures diversity and balanced coverage across users.

Overall, the experiments confirm that the proposed LLM-Guided Exemplar Selection substantially enhances few-shot HAR performance. By merging semantic reasoning, structural regularization, and ML gating, the system achieves high interpretability, balanced generalization, and robustness under limited labeled data. These results validate that semantic priors from LLMs can act as an effective knowledge layer in sensor-based recognition systems.
\section{Discussion}
\label{section:discussion}
The experimental results confirm that the proposed LLM-Guided Exemplar Selection framework consistently improves few-shot Human Activity Recognition (HAR) performance compared with classical methods such as Random Sampling, Herding, and K-Center. These improvements show that incorporating semantic reasoning from LLMs enhances not only accuracy but also the interpretability of sensor-based representations.

The framework’s strength lies in the interaction of four key components: LLM knowledge, PageRank-based structure, hubness penalty, and Facility Location optimization. LLM knowledge adds conceptual understanding of activities; PageRank maintains structural balance; hubness prevents bias; and Facility Location ensures diversity. The ablation results confirmed that removing any component reduces performance, especially Facility Location, underscoring the need to balance local and global representations.

The LLM serves not only as a semantic feature generator but also as a provider of structured knowledge, offering label-feature weights, inter-class confusability, and exemplar budget multipliers. Through these priors, the model interprets signals both numerically and conceptually, understanding, for instance, the semantic difference between walking upstairs and walking downstairs despite similar motion patterns.

The LLM's knowledge encoding has a positive effect on final performance. When this component is removed, the macro F1 score drops by about $(2.05)$ points. This shows that the LLM's semantic cues help the system avoid misleading or noisy exemplars, especially for naturally similar activity pairs. While the geometric and graph-based components describe how samples relate in feature space, the LLM prior adds an extra layer of understanding about how the activities themselves differ. This combination results in more reliable exemplar choices and better generalization when only a small amount of training data is available.

It is also important to highlight that the proposed framework maintains lightweight computational cost during inference. The runtime comparison in Appendix~\ref{time} shows that prediction speed is nearly identical to baseline methods across all six classifiers, indicating that the semantic–structural exemplar selection does not impose additional overhead and remains suitable for real-time or wearable-sensor environments.

Additionally, the Machine Learning Gate (ML Gate) enhances inference stability by separating static and dynamic activity domains before final prediction. This simple yet effective mechanism ensures that models like Logistic Regression, which achieved the best macro F1-score (88.79\%), can leverage LLM reasoning efficiently without complex architectures.

The proposed method is not limited to the UCI-HAR dataset. Because the framework relies only on basic statistical features and class-wise summaries, it can be applied to other wearable-sensor datasets, such as WISDM, PAMAP2, and RealWorld HAR. The LLM priors are also generated directly from the training data of each dataset, so the method can naturally adapt to different activity sets or sensor configurations. This makes the approach suitable for many real-world scenarios where labeled data are limited.

Although the framework shows strong performance, it also has several limitations. The quality of the LLM priors still depends on how well the model interprets the activity patterns, which may vary across datasets. The study also focuses on a single dataset, so evaluating the method on larger or noisier datasets would provide a clearer picture of its robustness. In addition, the current implementation uses simple statistical and PCA-based features; using richer time-series features may further improve results. Finally, while the LLM outputs are cached, the initial generation still requires access to an external API.

Overall, this study extends few-shot learning in HAR toward knowledge-guided reasoning, bridging data-driven and semantic approaches. The framework enables exemplar selection not only by geometric similarity but also by conceptual meaning, producing a system that is both accurate and explainable, valuable for real-world applications such as healthcare and activity monitoring, where data are scarce and labeling is costly.
\section{Conclusion}
\label{section:conclusion}
This research introduced an LLM-Guided Exemplar Selection framework that integrates semantic reasoning, structural regularization, and hybrid optimization for few-shot HAR. By combining LLM-derived features, graph-based scoring, and the ML Gate, the system achieved notable improvements over existing baselines, with an average macro F1-score gain of +10\% and peak performance of 88.79\% using Logistic Regression. Each component, semantic prior, PageRank, hubness control, and facility location, proved essential for maintaining stability and generalization.

Beyond performance, this approach establishes a new paradigm in sensor-based learning: using knowledge-guided reasoning to make the exemplar selection process semantically interpretable. It demonstrates that LLMs can serve as effective reasoning agents, enriching sensor representations and improving model robustness even with minimal labeled data.

Future work will explore multimodal fusion and lightweight or local LLM implementations for edge devices. These directions will further enhance adaptability, efficiency, and real-time applicability. In conclusion, the proposed framework bridges statistical learning and semantic reasoning, marking a step toward next-generation HAR systems that not only recognize activities but also understand their meaning.

\bibliographystyle{plain}
\bibliography{Authors/bibtex}
\clearpage 

\appendix 
\section{Appendix: LLM Prompting Design}
\label{prompt_design_app}
\begin{figure}[!ht]
\centering
\tcbset{colback=white, colframe=black, fonttitle=\bfseries}
\begin{tcolorbox}[title=LLM-based semantic feature prompting, breakable]
\footnotesize
\textbf{<SYSTEM>}: You are a sensor-HAR feature designer. Propose compact semantic features as linear combinations of the provided base channels. Return STRICT JSON only.\\
\textbf{<USER>}:\\
Base channels (standardized): acc\_x\_mean, acc\_x\_std, acc\_y\_mean, acc\_y\_std, acc\_z\_mean, acc\_z\_std, gyr\_x\_mean, gyr\_x\_std, gyr\_y\_mean, gyr\_y\_std, gyr\_z\_mean, gyr\_z\_std.\\
Class-wise mean table (train only): \{\texttt{train\_data}\}.\\
Design 6 discriminative SEMANTIC features (short lowercase names).\\
Each feature must be a LINEAR combination of the base channels with weights in [-2,2].\\
Return JSON ONLY like:\\
\{\\
  "features": [\\
    \{"name": "verticality", "weights": \{"acc\_z\_mean": 0.8, "acc\_z\_std": 0.6\}\},\\
    \{"name": "hip\_rotation", "weights": \{"gyr\_y\_std": 1.1, "gyr\_x\_std": 0.3\}\},\\
    ...\\
  ]\}
\end{tcolorbox}
\caption{Prompt design for LLM-based semantic feature generation.}
\label{fig:LLM-Semantic-Feature}
\end{figure}

\begin{figure}[!ht]
\centering
\tcbset{colback=white, colframe=black, fonttitle=\bfseries}
\begin{tcolorbox}[title=LLM Knowledge Representation Prompting, breakable]
\footnotesize
\textbf{<SYSTEM>}: You are a careful scientific assistant for smartphone inertial HAR. Return a SINGLE valid JSON object and nothing else. Do not use any test-set information. Use only the provided labels and features.\\
\textbf{<USER>}:\\
We need domain knowledge to guide exemplar selection.\\
Allowed labels: \{\texttt{label\_dataset}\}.\\
Allowed features (standardized): \{\texttt{features}\}.\\
Class-wise means over TRAIN (for context only): \{\texttt{train\_data}\}.\\
Return a JSON with keys:
\begin{itemize}
    \item "label\_feature\_weights": map label$\rightarrow$map feature$\rightarrow$weight in [-1.5, 1.5]
    \item "confusability": map labelA$\rightarrow$map labelB$\rightarrow$weight in [0, 1.2]
    \item "label\_budget\_multiplier": map label$\rightarrow$multiplier in [0.8, 1.3]
\end{itemize}
Rules:
\begin{itemize}
    \item Use ONLY the allowed labels and features.
    \item Keep JSON minimal. No comments, no trailing commas, no extra text.
\end{itemize}
\end{tcolorbox}
\caption{Prompt design for LLM knowledge representation.}
\label{fig:LLM-Know-Prompting}
\end{figure}

\section{Appendix: Pseudocode}
\label{pseudocode}
\subsection{Algorithm 1: Semantic Feature Generation}
\label{algo_b1}
\begin{algorithm}[H]
\caption{LLM-Guided Semantic Feature Synthesis}
\KwIn{$D_{\text{train}}$, $D_{\text{test}}$ (both standardized on $F_{\text{base}}$), $F_{\text{base}}$ (set of base features), desired count $N$ (default $=6$)}
\KwOut{$F_{\text{semantic}}$ (set of synthesized semantic features) columns appended to $D_{\text{train}}$ and $D_{\text{test}}$}

\BlankLine
\Comment{1) Compute class-wise statistics}
Compute class-wise mean table $M \leftarrow \text{Means}(D_{\text{train}})$\;

\BlankLine
\Comment{2) Compose and send the prompt to the LLM}
Form prompt $P$: “Design $N$ semantic features as linear combinations of $F_{\text{base}}$”\;
Send $P$ (with $F_{\text{base}}$ and $M$ as context) to LLM; receive JSON spec $S$\;
\Comment{$S.\texttt{features}$ is a list of \{\texttt{name}, \texttt{weights}\}}

\BlankLine
\Comment{3) Construct semantic features from JSON}
Initialize $F_{\text{semantic}} \leftarrow \varnothing$\;
\ForEach{\texttt{spec} $\in S.\texttt{features}$}{
    Let \texttt{spec.name} be $f$ and \texttt{spec.weights} be coefficients $\{\alpha_j\}$\;
    Define $f(x) \leftarrow \sum_{j} \alpha_{j}\,\text{base\_feature}_{j}(x)$\;
    Append $f$ to $F_{\text{semantic}}$\;
}

\BlankLine
\Return $F_{\text{semantic}}$\;
\end{algorithm}
\FloatBarrier
\subsection{Algorithm 2: LLM Knowledge Extraction}
\label{algo_b2}
\begin{algorithm}[H]
\caption{LLM-Guided Knowledge Induction (Feature--Label Relations \& Confusability)}
\KwIn{$D_{\text{train}}$ (labeled, standardized), $F_{\text{semantic}}$}
\KwOut{$K=\{\texttt{label\_feature\_weights},\ \texttt{confusability},\ \texttt{label\_budget\_multiplier}\}$}

\BlankLine
\Comment{1) Class-wise statistics over semantic features}
Compute class-wise means $M \leftarrow \text{Means}(D_{\text{train}}, F_{\text{semantic}})$\;

\BlankLine
\Comment{2) Compose and send structured prompt}
Form prompt $P$: “Generate knowledge JSON describing feature--label relations and confusion.”\;
Send $P$ with context $\{F_{\text{semantic}}, M\}$ to LLM; receive raw JSON $K_{\text{raw}}$\;

\BlankLine
\Comment{3) Parse \& schema validation}
Validate schema keys in $K_{\text{raw}}$: \{\texttt{label\_feature\_weights}, \texttt{confusability}, \texttt{label\_budget\_multiplier}\}\;
Coerce missing/invalid entries to defaults (e.g., zeros)\;

\BlankLine
\Comment{4) Constrain label--feature weights to $[-2,2]$}
\ForEach{label $\ell$}{
  \ForEach{feature $f \in F_{\text{semantic}}$}{
    $w \leftarrow K_{\text{raw}}[\texttt{label\_feature\_weights}][\ell][f]$\;
    $w \leftarrow \min\{2,\ \max\{-2,\ w\}\}$ \tcp*[r]{clamp to $[-2,2]$}
    $K[\texttt{label\_feature\_weights}][\ell][f] \leftarrow w$\;
  }
}

\BlankLine
\Comment{5) Normalize confusability to $[0,1.5]$}
\ForEach{pair $(\ell_a,\ell_b)$ with $\ell_a \neq \ell_b$}{
  $c \leftarrow K_{\text{raw}}[\texttt{confusability}][\ell_a][\ell_b]$\;
  \If{$c < 0$}{ $c \leftarrow 0$ }
  \If{$c > 1.5$}{ $c \leftarrow 1.5$ }
  $K[\texttt{confusability}][\ell_a][\ell_b] \leftarrow c$\;
}

\BlankLine
\Comment{6) Budget multiplier sanitization}
\ForEach{label $\ell$}{
  $b \leftarrow K_{\text{raw}}[\texttt{label\_budget\_multiplier}][\ell]$\;
  \If{$b$ is missing}{ $b \leftarrow 1.0$ }
  $b \leftarrow \min\{1.3,\ \max\{0.8,\ b\}\}$ \tcp*[r]{clamp to $[0.8,1.3]$}
  $K[\texttt{label\_budget\_multiplier}][\ell] \leftarrow $ $b$\;
}

\BlankLine
\Return $K$\;
\end{algorithm}
\FloatBarrier
\subsection{Algorithm 3: Hybrid Scoring}
\label{algo_b3}
\begin{algorithm}[H]
\caption{Hybrid Candidate Scoring (LLM Prior + Graph + Margin + Hubness)}
\KwIn{$X$ (candidate samples), $V$ (validation set), $K$ (knowledge JSON)}
\KwOut{$S(i)$ for each candidate $i \in X$}

\BlankLine
\Comment{0) Precompute structures from $X \cup V$}
Build mutual-$k$NN graph $G$ with $k{=}10$ on feature space; symmetrize edges\;
Compute PageRank scores $P(i)$ on $G$ with damping $\alpha$ (e.g., $0.85$)\;
Estimate hubness $H(i)$ (e.g., in-degree or top-$k$ reversed neighbor frequency)\;

\BlankLine
\Comment{1) Iterate candidates}
\ForEach{$i \in X$}{
  \Comment{(a) Positive/negative margin}
  Compute class scores on $V$ (or a local neighborhood) and derive margin\;
  $M(i) \leftarrow s_{+}(i) - s_{-}(i)$ \tcp*[r]{e.g., pos vs. nearest negative}

  \BlankLine
  \Comment{(b) Semantic prior from knowledge $K$}
  Derive $T(i)$ using $K$ (e.g., label--feature weights and confusability)\;
  \tcp{Example: $T(i)=\max_{\ell}\left(\sum_{f} w_{\ell,f}\,x_f(i)\right) - \gamma \cdot \max_{\ell'\neq \ell} \text{conf}(\ell,\ell')$}

  \BlankLine
  \Comment{(c) Combine with weights $\lambda_1,\lambda_2,\lambda_3,\lambda_4 \ge 0$}
  $S(i) \leftarrow \lambda_1\,M(i) + \lambda_2\,P(i) - \lambda_3\,H(i) + \lambda_4\,T(i)$\;
}

\BlankLine
\Return $S(i)$ for all $i \in X$\;
\end{algorithm}
\FloatBarrier
\subsection{Algorithm 4: Facility Location-based Selection}
\label{algo_b4}
\begin{algorithm}[H]
\caption{Global Exemplar Selection via Hybrid Scoring and Facility-Gain Optimization}
\KwIn{$X_{\text{train}}$ (training samples), $S(i)$ (hybrid scores), $L$ (class labels)}
\KwOut{$E$ (set of selected exemplars)}

\BlankLine
\Comment{1) Initialization}
Initialize $E \leftarrow \varnothing$\;
\ForEach{class $c \in L$}{
  Initialize class exemplar set $E_c \leftarrow \varnothing$\;
  Determine budget $b(c)$ based on prior (e.g., $K[\texttt{label\_budget\_multiplier}]$)\;
  
  \BlankLine
  \Comment{2) Candidate pool and selection loop}
  $C_c \leftarrow \{i \in X_{\text{train}} \mid \text{label}(i)=c\}$\;
  \While{$|E_c| < b(c)$}{
     \Comment{Select exemplar maximizing facility-gain and hybrid score}
     $j^\star \leftarrow \arg\max_{j \in C_c \setminus E_c} \big[\Delta F(j \mid E_c) + \alpha \, S(j)\big]$\;
     $E_c \leftarrow E_c \cup \{j^\star\}$\;
  }
}

\BlankLine
\Comment{3) Combine exemplars from all classes}
$E \leftarrow \bigcup_{c \in L} E_c$\;

\BlankLine
\Return $E$\;
\end{algorithm}
\FloatBarrier
\subsection{Algorithm 5: Main Procedure}
\label{algo_b5}
\begin{algorithm}[H]
\caption{End-to-End Evaluation Pipeline with ML Gate}
\KwIn{$D_{\text{train}}$ (raw), $D_{\text{val}}$ (raw), $D_{\text{test}}$ (raw), threshold $\tau$ (default $=0.50$)}
\KwOut{Final metrics: Macro-F1}

\BlankLine
\Comment{1) Preprocessing \& standardization (fit on train only)}
Fit standardizer $\mathcal{S}$ on $D_{\text{train}}$ base features;\\
$F^{\text{base}}_{\text{train}} \leftarrow \mathcal{S}(D_{\text{train}})$,\quad
$F^{\text{base}}_{\text{val}}   \leftarrow \mathcal{S}(D_{\text{val}})$,\quad
$F^{\text{base}}_{\text{test}}  \leftarrow \mathcal{S}(D_{\text{test}})$\;

\BlankLine
\Comment{2) Semantic features (Algorithm~\ref{algo_b1})}
$F_{\text{semantic}} \leftarrow \textsc{LLM\_SemanticFeatures}(D_{\text{train}}, F^{\text{base}}_{\text{train}})$\;

\BlankLine
\Comment{3) Knowledge JSON (Algorithm~\ref{algo_b2})}
$K \leftarrow \textsc{LLM\_KnowledgeJSON}(D_{\text{train}}, F_{\text{semantic}})$\;

\BlankLine
\Comment{4) Combined feature space (fit PCA/Spectral on train only)}
Learn PCA on $F^{\text{base}}_{\text{train}}$; obtain $\text{PCA}_{\text{train/val/test}}$\;
Learn Spectral-PCA on $F^{\text{base}}_{\text{train}}$; obtain $\text{SpecPCA}_{\text{train/val/test}}$\;
Compose
$F^{\text{all}}_{\text{train}} \leftarrow \{F^{\text{base}}_{\text{train}}, F_{\text{semantic}}, \text{PCA}_{\text{train}}, \text{SpecPCA}_{\text{train}}\}$,\\
$F^{\text{all}}_{\text{val}}   \leftarrow \{F^{\text{base}}_{\text{val}},   \text{PCA}_{\text{val}},   \text{SpecPCA}_{\text{val}}\}$,\\
$F^{\text{all}}_{\text{test}}  \leftarrow \{F^{\text{base}}_{\text{test}},  \text{PCA}_{\text{test}},  \text{SpecPCA}_{\text{test}}\}$\;

\BlankLine
\Comment{4.5) Train ML Gate (STATIC vs DYNAMIC)}
Define $\mathcal{L}_{\text{STATIC}}{=}\{\text{LAYING,SITTING,STANDING}\}$, 
$\mathcal{L}_{\text{DYN}}{=}\{\text{WALKING,WALKING\_UPSTAIRS,WALKING\_DOWNSTAIRS}\}$\;
Build $y_{\text{gate}}$ on $D_{\text{train}}$: $1$ if label $\in \mathcal{L}_{\text{DYN}}$ else $0$\;
Train logistic gate $\mathcal{G}$ on $F^{\text{all}}_{\text{train}}$ (class\_weight=balanced, max\_iter=800)\;

\BlankLine
\Comment{5) Hybrid scoring for candidate selection (Algorithm~\ref{algo_b3})}
$S(i) \leftarrow \textsc{HybridScore}(X_{\text{train}}{:=}F^{\text{all}}_{\text{train}},\ V{:=}F^{\text{all}}_{\text{val}},\ K)$\;

\BlankLine
\Comment{6) Exemplar selection (Algorithm~\ref{algo_b4})}
$E \leftarrow \textsc{SelectExemplars}(X_{\text{train}}, S(i), L_{\text{train}})$\;

\BlankLine
\Comment{7) Train ML models on selected exemplars}
Train $\mathcal{M}_{\text{KNN}}, \mathcal{M}_{\text{LogReg}}, \mathcal{M}_{\text{RF}}, \ldots$ on $(E,\ \text{labels}(E))$\;

\BlankLine
\Comment{8) Evaluate on test set with ML Gate}
\ForEach{$x \in F^{\text{all}}_{\text{test}}$}{
  $p_{\text{dyn}} \leftarrow \mathcal{G}.\mathrm{predict\_proba}(x)[\mathrm{DYNAMIC}]$\;
  $\mathcal{A}(x) \leftarrow \big(\mathcal{L}_{\text{DYN}} \ \textbf{if}\ p_{\text{dyn}}\ge\tau\ \textbf{else}\ \mathcal{L}_{\text{STATIC}}\big)$\;
  \ForEach{$\mathcal{M} \in \{\mathcal{M}_{\text{KNN}}, \mathcal{M}_{\text{LogReg}}, \mathcal{M}_{\text{RF}}, \ldots\}$}{
     Get raw $q_{\mathcal{M}}(\ell\mid x)$; set $d_{\mathcal{M}}(\ell){:=}q_{\mathcal{M}}(\ell\mid x)$ if $\ell\in\mathcal{A}(x)$ else $0$\;
     $s{:=}\sum_{\ell\in\mathcal{A}(x)} d_{\mathcal{M}}(\ell)$; 
     \uIf{$s \le 0$}{ $d_{\mathcal{M}}(\ell){:=}1/|\mathcal{A}(x)|$ for $\ell\in\mathcal{A}(x)$ } 
     \Else{ $d_{\mathcal{M}}(\ell){:=}d_{\mathcal{M}}(\ell)/s$ for $\ell\in\mathcal{A}(x)$ }
     $\hat{y}_{\mathcal{M}}(x) \leftarrow \arg\max_{\ell\in\mathcal{A}(x)} d_{\mathcal{M}}(\ell)$\;
  }
}
Compute Macro-F1 over $(\hat{y}_{\mathcal{M}}, y_{\text{test}})$\;

\BlankLine
\Return $\{\text{Macro-F1}\}$\;
\end{algorithm}

\section{Appendix: LLM Response}
\label{llm_response}
\subsection{LLM Response for semantic feature}
\label{semantic}
\begin{figure}[!ht]
\centering
\footnotesize
\begin{minipage}{0.9\linewidth}
\hrule\vspace{3pt}
\textbf{LLM response for semantic feature}
\vspace{3pt}\hrule
\begin{verbatim}
{
  "features": [
    {
      "name": "verticality",
      "weights": {
        "acc_z_mean": 0.8,
        "acc_z_std": 0.6
      }
    },
    {
      "name": "lateral_movement",
      "weights": {
        "acc_x_mean": 1.2,
        "acc_y_mean": -0.5
      }
    },
    {
      "name": "rotation",
      "weights": {
        "gyr_z_mean": 1.5,
        "gyr_z_std": 0.5
      }
    },
    {
      "name": "acceleration_magnitude",
      "weights": {
        "acc_x_mean": 0.5,
        "acc_y_mean": 0.5,
        "acc_z_mean": 0.5
      }
    },
    {
      "name": "gyr_std_dev",
      "weights": {
        "gyr_x_std": 1.0,
        "gyr_y_std": 1.0,
        "gyr_z_std": 1.0
      }
    },
    {
      "name": "activity_level",
      "weights": {
        "acc_x_std": 1.0,
        "acc_y_std": 1.0,
        "acc_z_std": 1.0
      }
    }
  ]
}
\end{verbatim}
\vspace{3pt}\hrule
\caption{LLM response for semantic feature synthesis prompt.}
\label{fig:LLM-Semantic-Feature-response}
\end{minipage}
\end{figure}
This appendix illustrates how the Large Language Model (LLM) generates new semantic features from the existing statistical features of accelerometer and gyroscope data, as illustrated in Fig. \ref{fig:LLM-Semantic-Feature-response}. The LLM receives a summary of channel means and standard deviations for each activity class and is instructed to design meaningful linear combinations that represent interpretable motion concepts. 
As a result, six new semantic features are produced: \textit{verticality}, \textit{lateral movement}, \textit{rotation}, \textit{acceleration magnitude}, \textit{gyroscope standard deviation}, and \textit{activity level}.

Each semantic feature is computed as a weighted sum of the existing statistical features, with the LLM suggesting weights in the range $[-2, 2]$. For example, the feature \textit{verticality} combines the mean and standard deviation of the $z$-axis acceleration to represent the intensity of vertical body motion:
\[
\text{Verticality} = 0.8 \times acc\_z\_mean + 0.6 \times acc\_z\_std.
\]
If $acc\_z\_mean=0.45$ and $acc\_z\_std=0.30$, the resulting value is $0.54$, indicating a relatively strong vertical component, such as when walking upstairs or standing. Similarly, the \textit{rotation} feature is formed from the $z$-axis gyroscope values:
\[
\text{Rotation} = 1.5 \times gyr\_z\_mean + 0.5 \times gyr\_z\_std,
\]
which reflects rotational intensity during dynamic movements. 
The \textit{activity level} feature aggregates the standard deviations of the accelerometer channels,
\[
\text{Activity Level} = acc\_x\_std + acc\_y\_std + acc\_z\_std,
\]
providing a measure of overall movement variability, where higher values correspond to more active behavior.

Table~\ref{tab:appendix_semfeat} summarizes all six semantic features and their interpretations.
\begin{table}[H]
\centering
\caption{Example of semantic features generated by the LLM.}
\label{tab:appendix_semfeat}
\renewcommand{\arraystretch}{1.2}
\setlength{\tabcolsep}{5pt}
\begin{tabularx}{\linewidth}{
>{\raggedright\arraybackslash}X
>{\centering\arraybackslash}X
>{\raggedright\arraybackslash}X
}
\toprule
\textbf{Feature Name} & \textbf{Formula} & \textbf{Interpretation} \\
\midrule
Verticality & $0.8\,acc\_z\_mean + 0.6\,acc\_z\_std$ & Represents the vertical component of body movement. \\
Lateral Movement & $1.2\,acc\_x\_mean - 0.5\,acc\_y\_mean$ & Captures sideward or directional motion changes. \\
Rotation & $1.5\,gyr\_z\_mean + 0.5\,gyr\_z\_std$ & Measures rotational intensity of the torso or hips. \\
Acceleration Magnitude & $0.5(acc\_x\_mean + acc\_y\_mean + acc\_z\_mean)$ & Reflects overall acceleration strength. \\
Gyroscope Std. Dev. & $gyr\_x\_std + gyr\_y\_std + gyr\_z\_std$ & Indicates rotational variability across all axes. \\
Activity Level & $acc\_x\_std + acc\_y\_std + acc\_z\_std$ & Quantifies the overall intensity of body movement. \\
\bottomrule
\end{tabularx}
\end{table}

\begin{table}[H]
\centering
\caption{Comparison of Macro F1-scores (\%) across models with and without LLM-based semantic features. The \textit{Improvement} column shows the change in performance after adding semantic features.}
\label{tab:semantic_comparison}
\renewcommand{\arraystretch}{1.2}
\setlength{\tabcolsep}{5pt}
\begin{tabularx}{\linewidth}{
>{\raggedright\arraybackslash}X
>{\centering\arraybackslash}X
>{\centering\arraybackslash}X
>{\centering\arraybackslash}X
}
\toprule
\textbf{Model} &\textbf{Without Semantic Features} &\textbf{With Semantic Features} &\textbf{Improvement}\\
\midrule
kNN & 83.09 & \textbf{85.88} & +2.79\\
Logistic Regression & 87.22 &\textbf{88.79} & +1.56\\
Random Forest & \textbf{83.16} & 79.61 & -3.55\\
HistGradient Boosting & 16.99 & 16.99 & +0\\
Linear SVC & 71.51 & \textbf{72.85} & +1.34\\
Gaussian Naïve Bayes & 46.61 & \textbf{58.03} & +11.43\\
\midrule
\textbf{Mean} & 64.76 &\textbf{67.02} &\textbf{+2.26}\\
\bottomrule
\end{tabularx}
\end{table}
The quantitative comparison in Table~\ref{tab:semantic_comparison} shows that the inclusion of semantic features improved the overall mean Macro F1-score from 64.76\% to 67.02\%, representing a relative gain of +2.26\%. The most significant improvements were observed in k-Nearest Neighbors (+2.8 points) and Gaussian Naïve Bayes (+11.4 points), indicating that the semantic abstraction particularly benefits distance-based and probabilistic models by enhancing the expressiveness of the feature space. Logistic Regression also achieved a steady increase (+1.6 points), suggesting that the LLM-generated features align well with linear decision boundaries. A slight decrease was observed in Random Forest (-3.6 points), likely due to feature redundancy introduced by semantic composition, while HistGradient Boosting remained unchanged under few-shot constraints. Overall, these results confirm that adding semantic features improves model generalization and interpretability under limited data conditions.

In summary, this process allows the model to translate raw sensor data into semantically meaningful features. Rather than relying solely on geometric proximity or statistical similarity, these LLM-derived features enrich the representation space by introducing conceptual dimensions such as directionality, intensity, and rotation. This semantic enhancement improves data interpretability and enhances robust few-shot learning performance in Human Activity Recognition.

\subsection{LLM Response for LLM Knowledge (priors)}
\label{prior}
\begin{figure}[!ht]
\centering
\tiny
\begin{adjustbox}{max width=\linewidth}
\begin{minipage}{1.0\linewidth}
\hrule\vspace{2pt}
\textbf{LLM response for LLM Knowledge (priors)}
\vspace{3pt}\hrule
\begin{verbatim}
{
  "label_feature_weights": {
    "WALKING": {"acc_x_mean": 0.5, "acc_y_mean": -0.5, "acc_z_mean": -0.5, "gyr_x_std": 0.5, "gyr_y_std": 0.5
    },
    "WALKING_UPSTAIRS": {"acc_x_mean": 0.4, "acc_y_mean": -0.8, "acc_z_mean": -0.7, "gyr_x_mean": 0.3, "gyr_y_std": 0.5
    },
    "WALKING_DOWNSTAIRS": {"acc_x_mean": 0.5, "acc_y_mean": -0.5, "acc_z_mean": -0.4, "gyr_x_std": -0.3, "gyr_y_std": 0.5
    },
    "SITTING": {"acc_x_mean": 0.4, "acc_y_mean": 0.3, "acc_z_mean": 0.2, "gyr_x_std": -0.5, "gyr_y_std": -0.5
    },
    "STANDING": {"acc_x_mean": 0.5, "acc_y_mean": -0.5, "acc_z_mean": -0.3, "gyr_x_std": 0.2, "gyr_y_std": -0.5
    },
    "LAYING": {"acc_x_mean": -0.5, "acc_y_mean": 0.5, "acc_z_mean": 0.4, "gyr_x_std": -0.5, "gyr_y_std": -0.5
    }
  },
  "confusability": {
    "WALKING": {"WALKING_UPSTAIRS": 0.5, "WALKING_DOWNSTAIRS": 0.5, "SITTING": 0.3, "STANDING": 0.4
    },
    "WALKING_UPSTAIRS": {"WALKING": 0.5, "WALKING_DOWNSTAIRS": 0.6, "SITTING": 0.4, "STANDING": 0.5
    },
    "WALKING_DOWNSTAIRS": {"WALKING": 0.5, "WALKING_UPSTAIRS": 0.6, "SITTING": 0.4, "STANDING": 0.5
    },
    "SITTING": { "WALKING": 0.3, "WALKING_UPSTAIRS": 0.4, "WALKING_DOWNSTAIRS": 0.4, "STANDING": 0.5
    },
    "STANDING": {"WALKING": 0.4, "WALKING_UPSTAIRS": 0.5, "WALKING_DOWNSTAIRS": 0.5, "SITTING": 0.5
    },
    "LAYING": { "SITTING": 0.3, "STANDING": 0.4
    }
  },
  "label_budget_multiplier": {
    "WALKING": 1.2,
    "WALKING_UPSTAIRS": 1.1,
    "WALKING_DOWNSTAIRS": 1.1,
    "SITTING": 1.0,
    "STANDING": 1.0,
    "LAYING": 0.9
  }
}
\end{verbatim}
\vspace{3pt}\hrule
\caption{LLM Knowledge (priors): feature weights per label, confusability, and label-wise budget multipliers.}
\label{fig:llm-knowledge}
\end{minipage}
\end{adjustbox}
\vspace{4pt} 
\FloatBarrier
\end{figure}

The Large Language Model (LLM) was also prompted to generate structured knowledge that describes the semantic relationships among activities and their associated sensor features as illustrated in Fig. \ref{fig:llm-knowledge}. This knowledge is returned in JSON format and consists of four components: (1) \textit{label-feature weights}, (2) \textit{class confusability}, and (3) \textit{label budget multipliers}. Each component provides additional context that guides the exemplar selection process beyond raw geometric similarity.

\textbf{1) Label-Feature Weights.} This part assigns a numerical importance score to each feature for every activity class. For example, for the class \textit{WALKING}, the LLM assigns higher positive weights to the standard deviation of gyroscope signals (gyr\_x\_std, gyr\_y\_std) and moderate positive weights to acc\_x\_mean, indicating that walking involves rhythmic lateral movement and moderate body rotation. Meanwhile, \textit{WALKING\_UPSTAIRS} shows stronger influence from vertical acceleration (acc\_z\_mean=-0.7) and the mean gyroscope rate (gyr\_x\_mean=0.3), representing the upward motion pattern. For static activities such as \textit{SITTING} and \textit{LAYING}, negative weights are assigned to gyroscope variation, indicating minimal rotation and low movement intensity. This mapping helps the system compute a \textit{textual margin score} that measures how well each sample aligns semantically with its target class.

\textbf{2) Class Confusability.} This component quantifies the degree of similarity between classes, with values ranging from 0 (distinct) to 1 (highly similar). As expected, activities such as \textit{WALKING}, \textit{WALKING\_UPSTAIRS}, and \textit{WALKING\_DOWNSTAIRS} exhibit high mutual confusability values ($0.5$–$0.6$), reflecting their overlapping motion patterns. In contrast, static–dynamic pairs such as \textit{WALKING} vs. \textit{SITTING} have lower confusability ($0.3$–$0.4$), indicating distinct kinematic signatures. This matrix is used during exemplar scoring to weight negative samples proportionally to their semantic proximity, thus maintaining a realistic notion of inter-class difficulty.

\textbf{3) Label Budget Multiplier.} Finally, this component controls the number of exemplars allocated per class based on activity characteristics. 
Dynamic activities (\textit{WALKING}, \textit{WALKING\_UPSTAIRS}, \textit{WALKING\_DOWNSTAIRS}) receive slightly higher multipliers ($1.1$–$1.2$), allowing more exemplars to represent their complex and variable movements. In contrast, static classes such as \textit{SITTING}, \textit{STANDING}, and \textit{LAYING} are assigned lower or baseline multipliers ($0.9$–$1.0$), reflecting their lower intra-class variability.

In summary, this LLM-generated knowledge prior functions as a structured semantic guide that complements numerical feature learning. It encodes domain understanding of movement patterns, inter-class similarity, and class-specific variability, thereby allowing the exemplar selection framework to operate not only on geometric closeness but also on conceptual relevance.

\section{Appendix: Confusion Metric}
\label{confusion}
\begin{figure}[!ht]
\centering
\captionsetup{justification=centering}
\setlength{\tabcolsep}{1pt}
\begin{tabular}{cc}
\includegraphics[width=0.48\linewidth, keepaspectratio]{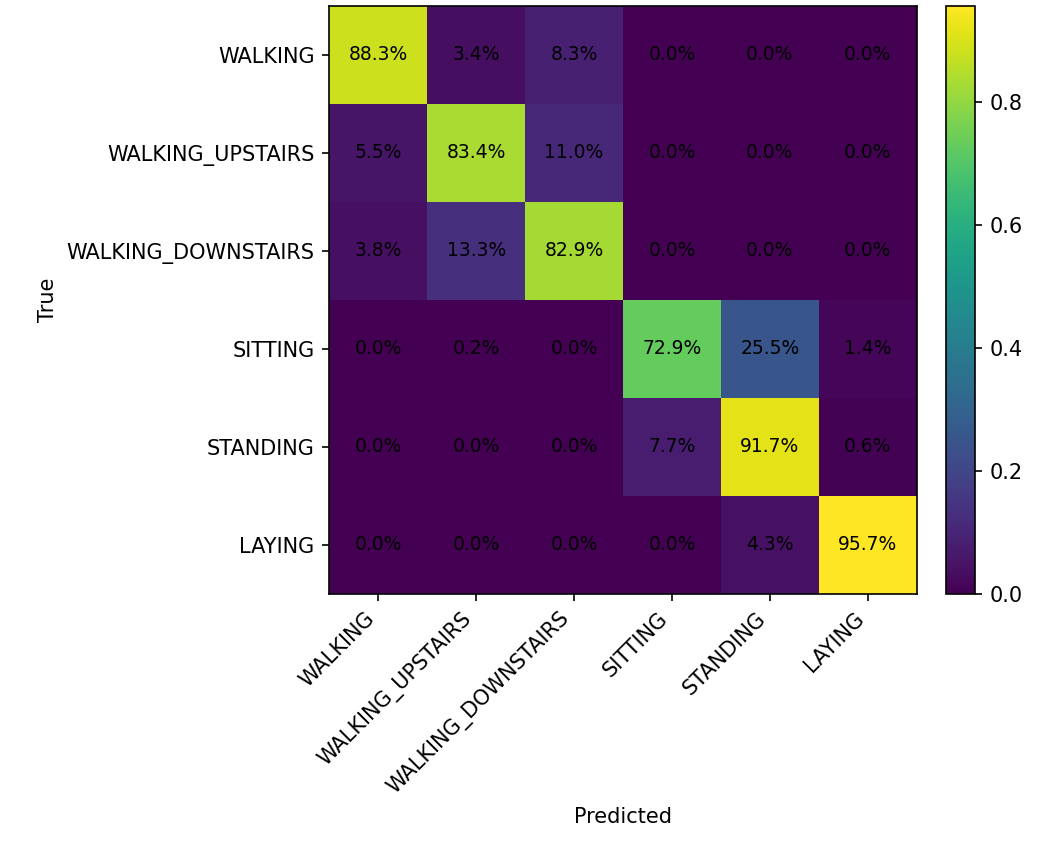} &
\includegraphics[width=0.48\linewidth, keepaspectratio]{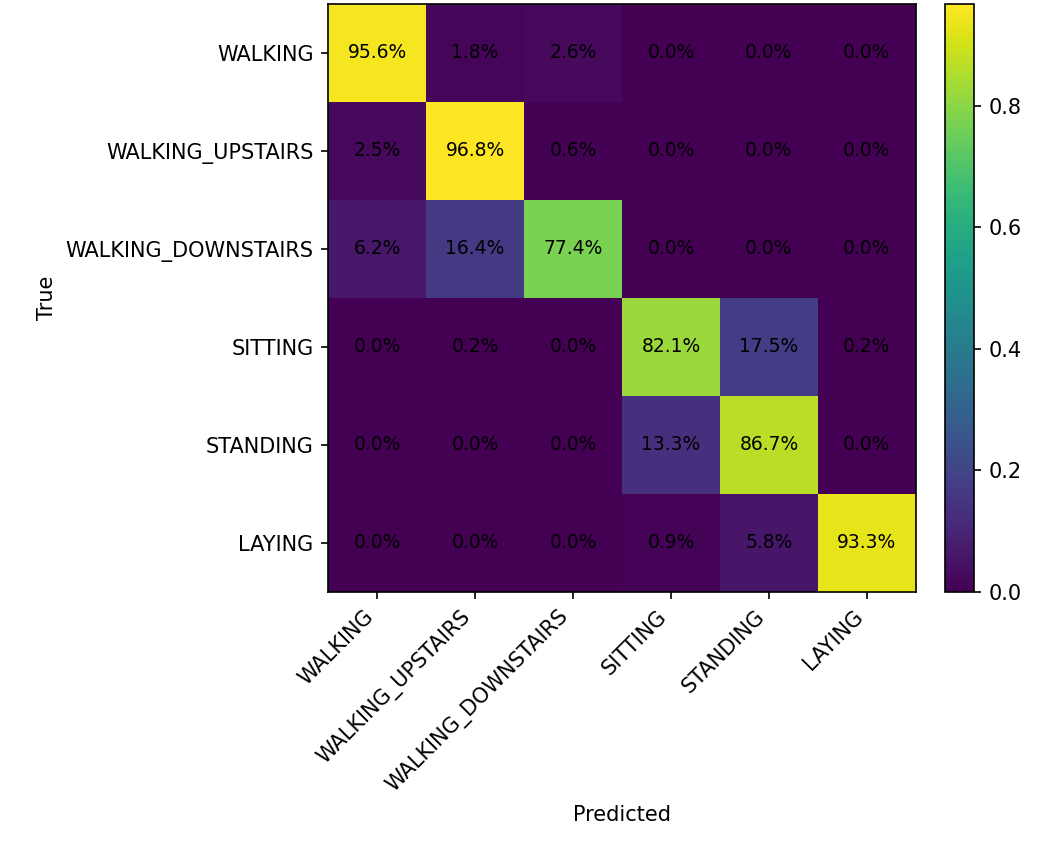} \\[-1pt]
(a) k-Nearest Neighbors (k-NN) & (b) Logistic Regression \\[4pt]
\includegraphics[width=0.48\linewidth, keepaspectratio]{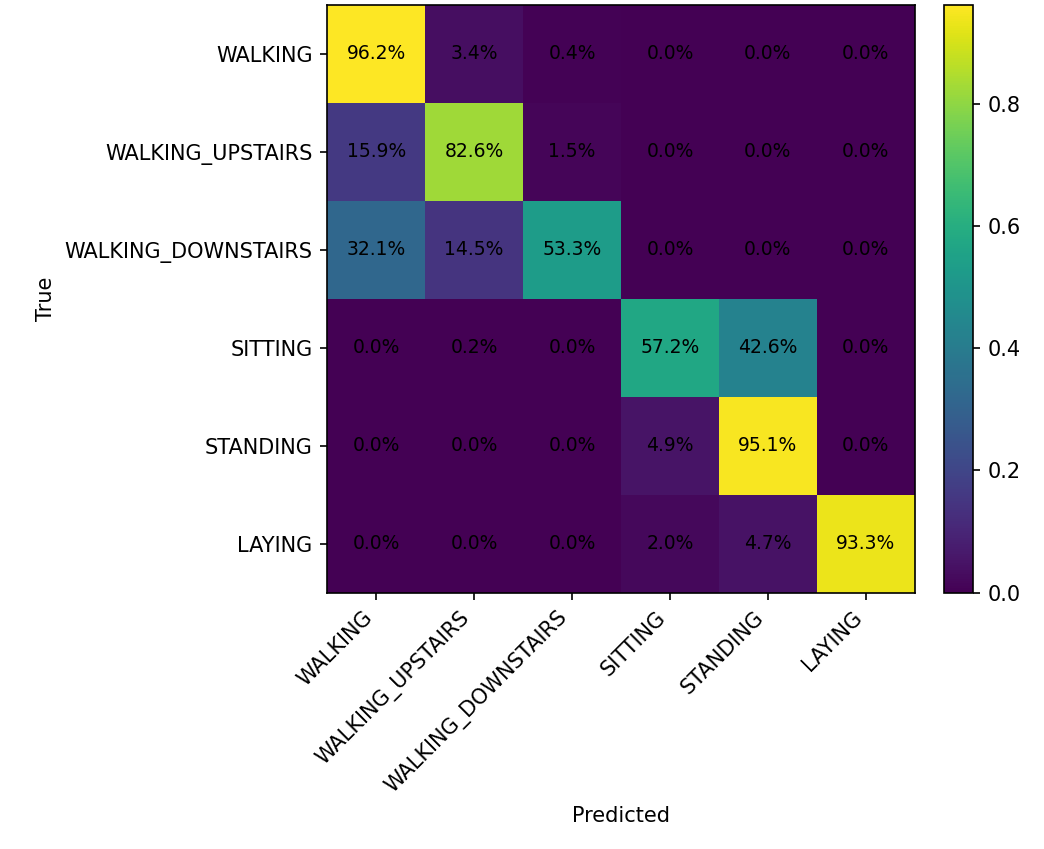} &
\includegraphics[width=0.48\linewidth, keepaspectratio]{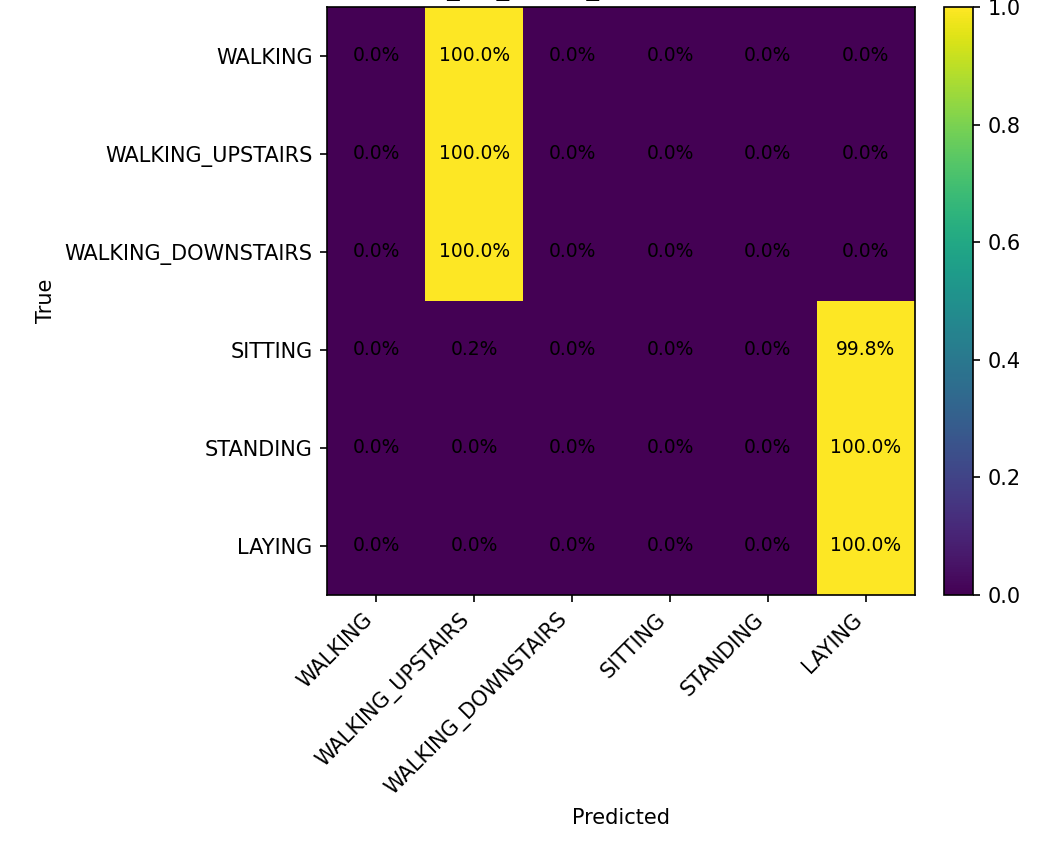}\\[-1pt]
(c) Random Forest & (d) Histogram Gradient Boosting\\[4pt]
\includegraphics[width=0.48\linewidth, keepaspectratio]{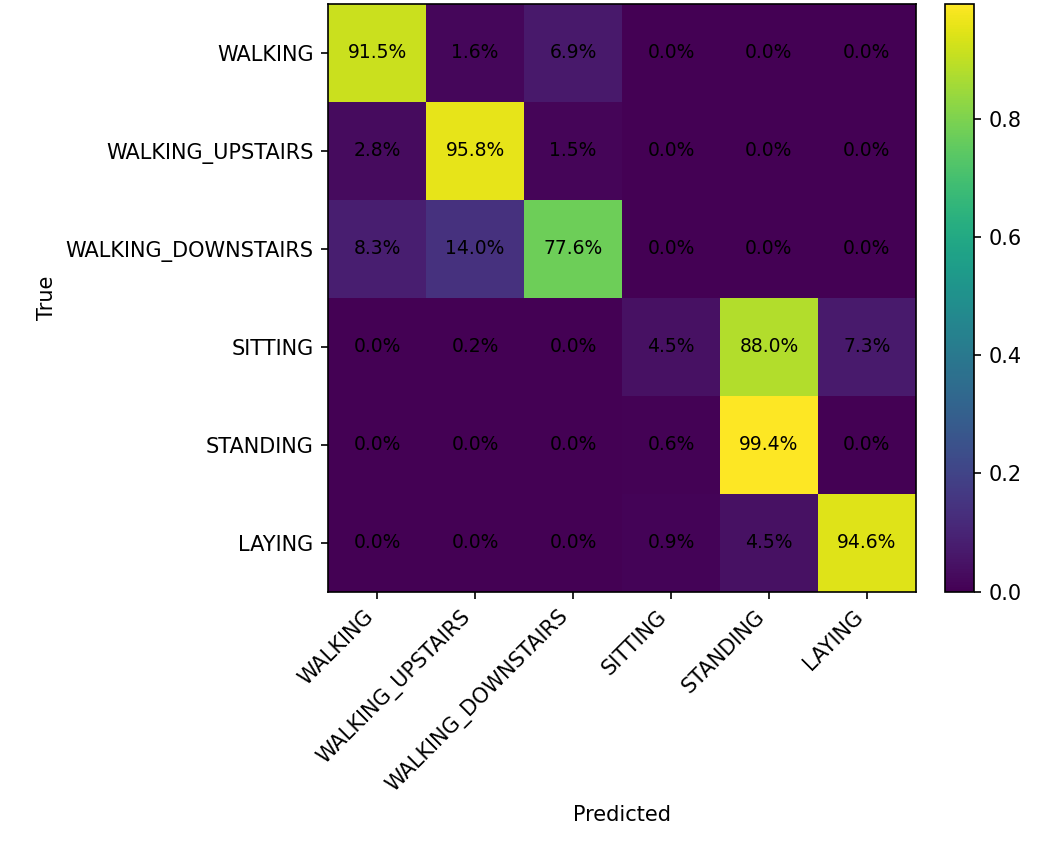} &
\includegraphics[width=0.48\linewidth, keepaspectratio]{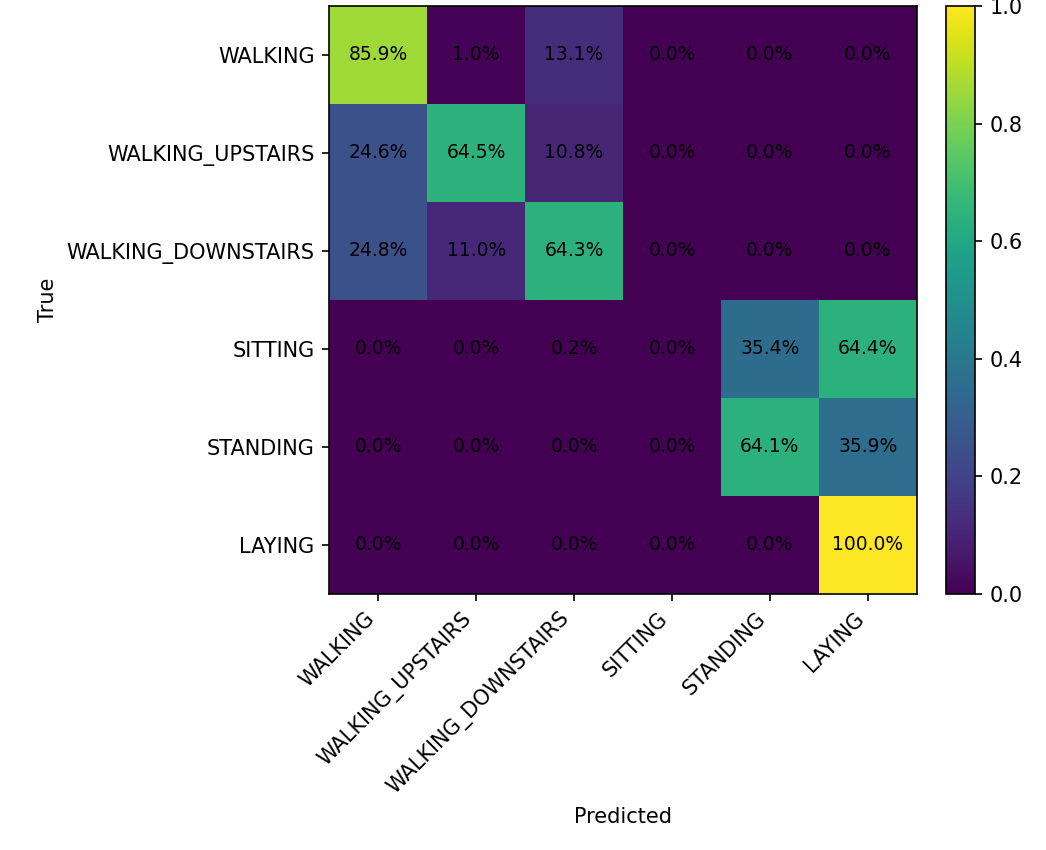} \\[-1pt]
(e) Linear SVC & (f) Gaussian Naïve Bayes
\end{tabular}
\caption{Confusion matrices of our proposed approach across six baseline classifiers. Each confusion matrix illustrates per-class prediction accuracy and misclassification patterns in the few-shot scenario.}
\label{fig:conf_matrix_ours_all}
\end{figure}

\section{Appendix: Effect of Machine Learning Gate (ML Gate)}
\label{mlgate}
Table~\ref{tab:mlgate_comparison} presents a quantitative comparison of model performance with and without the Machine Learning Gate (ML Gate). Across six classical classifiers, the inclusion of the ML Gate improved the mean macro F1 from 59.19\% to 67.02\% (+7.83 points).
The most pronounced gains were observed in the Gaussian Naïve Bayes (+29 points) and HistGradient Boosting (+12 points) models, where label space restriction helped reduce cross-category confusion. Moderate improvements were also noted in Linear SVC (+5 points). In contrast, distance-based and linear models (kNN, Logistic Regression) remained stably high. This indicates that the ML Gate mainly benefits models that are sensitive to overlapping feature distributions or imbalanced class priors.

Overall, the ML Gate enhances robustness by constraining predictions to either \textit{STATIC} or \textit{DYNAMIC} activity groups, thereby reducing semantic ambiguity during inference. It acts as a lightweight semantic filter that yields more consistent predictions across subjects without increasing computational cost.

\begin{table}[H]
\centering
\caption{Performance comparison with and without ML Gate (Macro F1, \%).}
\label{tab:mlgate_comparison}
\renewcommand{\arraystretch}{1.2}
\setlength{\tabcolsep}{6pt}
\begin{tabularx}{\linewidth}{
>{\raggedright\arraybackslash}X
>{\centering\arraybackslash}X
>{\centering\arraybackslash}X
>{\centering\arraybackslash}X
}
\toprule
\textbf{Model} & \textbf{Without ML Gate} & \textbf{With ML Gate} & \textbf{Improvement} \\
\midrule
kNN (Cosine) & 85.75 & \textbf{85.88} & +0.13 \\
Logistic Regression & 88.62 & \textbf{88.79} & +0.17 \\
Random Forest & 79.48 & \textbf{79.61} & +0.13 \\
HistGradient Boosting & 5.14 & \textbf{16.99} & +11.85 \\
Linear SVC & 67.48 & \textbf{72.85} & +5.37 \\
Gaussian Naïve Bayes & 28.69 & \textbf{58.03} & +29.35 \\
\midrule
\textbf{Mean} & 59.19 & \textbf{67.02} & +7.83 \\
\bottomrule
\end{tabularx}
\end{table}

\section{Appendix: Visualization of Exemplar Distribution}
\label{visualization}
This appendix provides additional qualitative results showing how the proposed LLM-guided exemplar selection affects the spatial distribution of training samples in the latent feature space. The visualizations use two-dimensional UMAP projections based on the full feature representation. Exemplars selected by different methods are overlaid on the same projection to illustrate how LLM knowledge influences sample placement.

\subsection{Global UMAP Projection}
\label{global_umap}
\begin{figure}[!ht]
\centering
\includegraphics[width=0.9\linewidth]{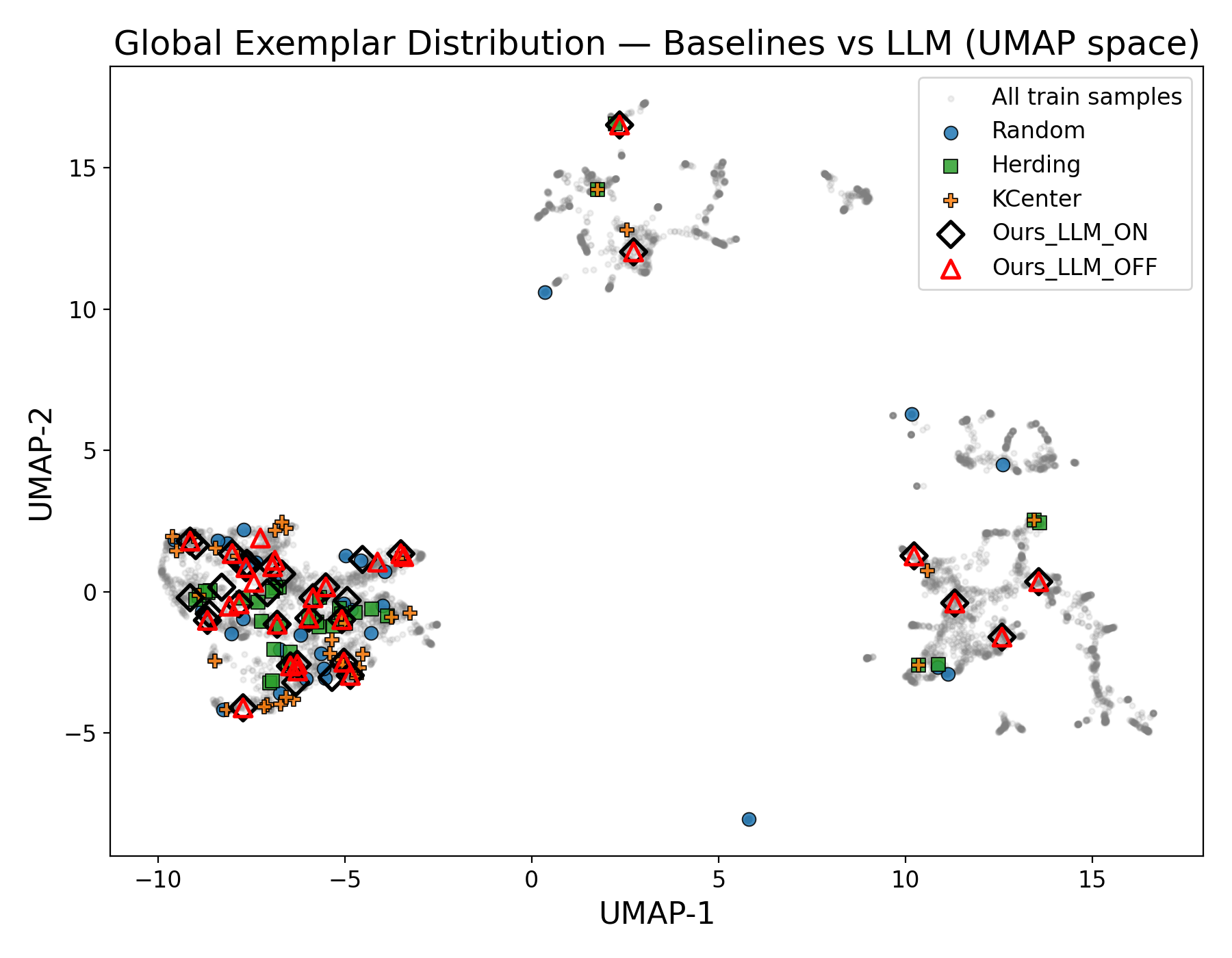}
\caption{Global UMAP projection of the training set with exemplars selected by different methods.}
\label{fig:umap-global-llm}
\end{figure}

Figure~\ref{fig:umap-global-llm} shows a global UMAP visualization of all training samples (gray), along with exemplars selected by Random, Herding, k-Center, Ours (LLM-OFF), and Ours (LLM-ON).  
Key observations:
\begin{itemize}
    \item Baseline methods either oversample the cluster center (Herding), distribute uniformly (k-Center), or scatter randomly.
    \item Ours (LLM-OFF) selects more informative samples than baselines due to margin- and graph-based scoring.
    \item Ours (LLM-ON) places exemplars in semantically meaningful subregions, influenced by label-feature weights, confusability, and LLM-informed budget allocation.
\end{itemize}
This demonstrates how LLM priors guide exemplar selection beyond geometric similarity.

\FloatBarrier

\subsection{Per-Class UMAP Visualization}
\label{umap_perclass}
\begin{figure}[H]
\centering
\captionsetup{justification=centering}
\setlength{\tabcolsep}{1pt}
\begin{tabular}{cc}
\includegraphics[width=0.5\linewidth, keepaspectratio]{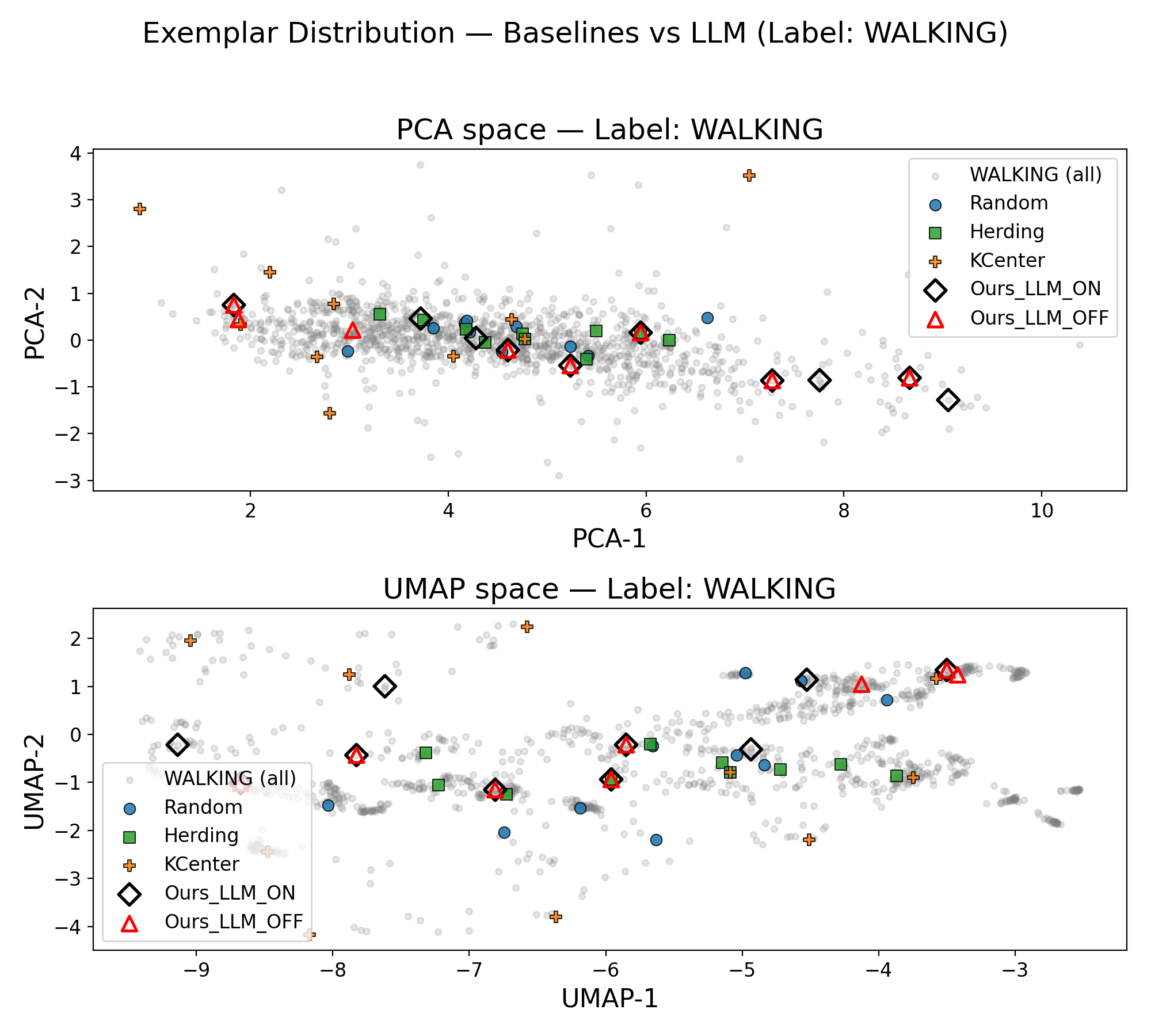} &
\includegraphics[width=0.5\linewidth, keepaspectratio]{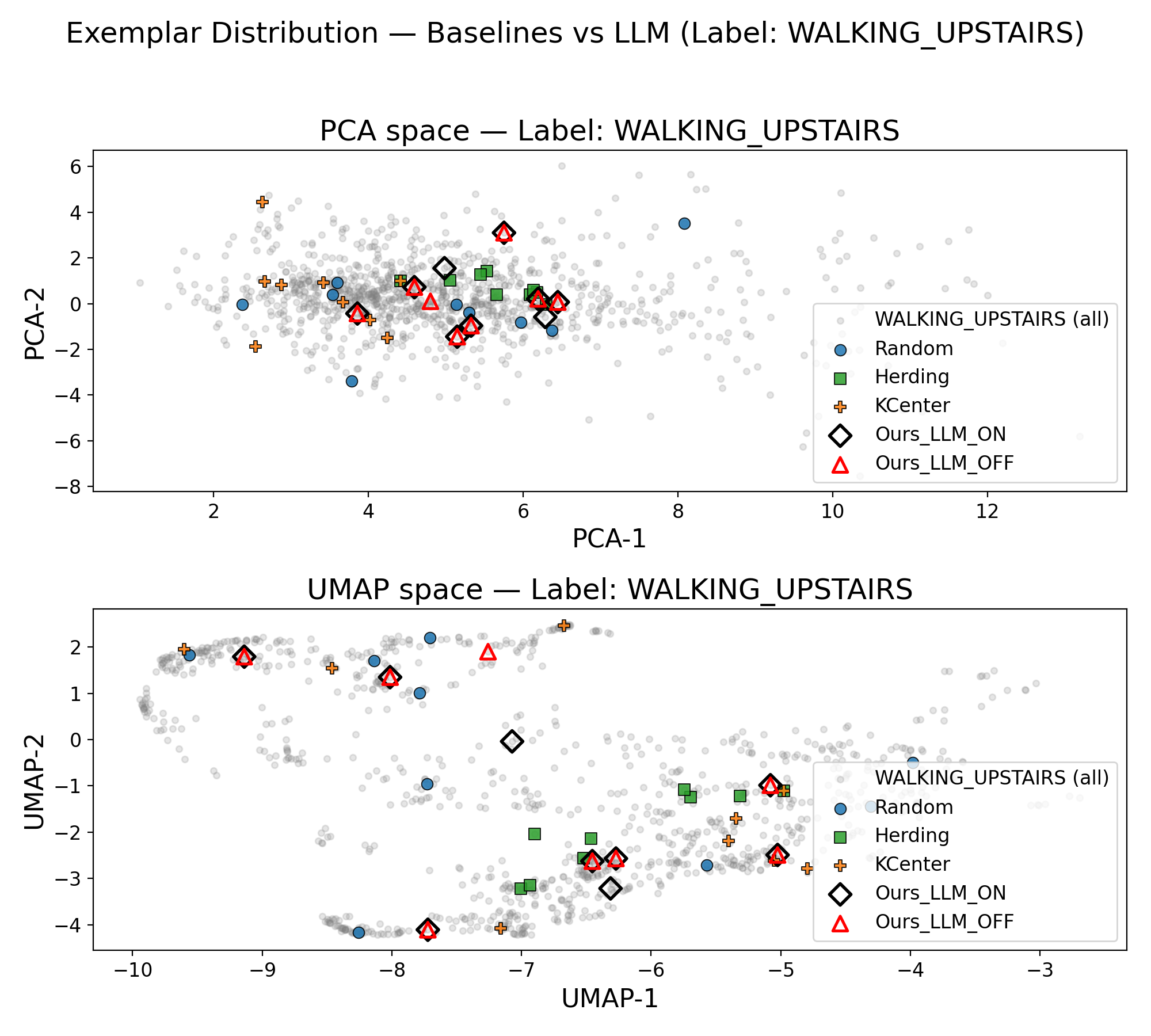} \\[-1pt]
(a) Walking & (b) Walking Upstairs \\[4pt]
\includegraphics[width=0.5\linewidth, keepaspectratio]{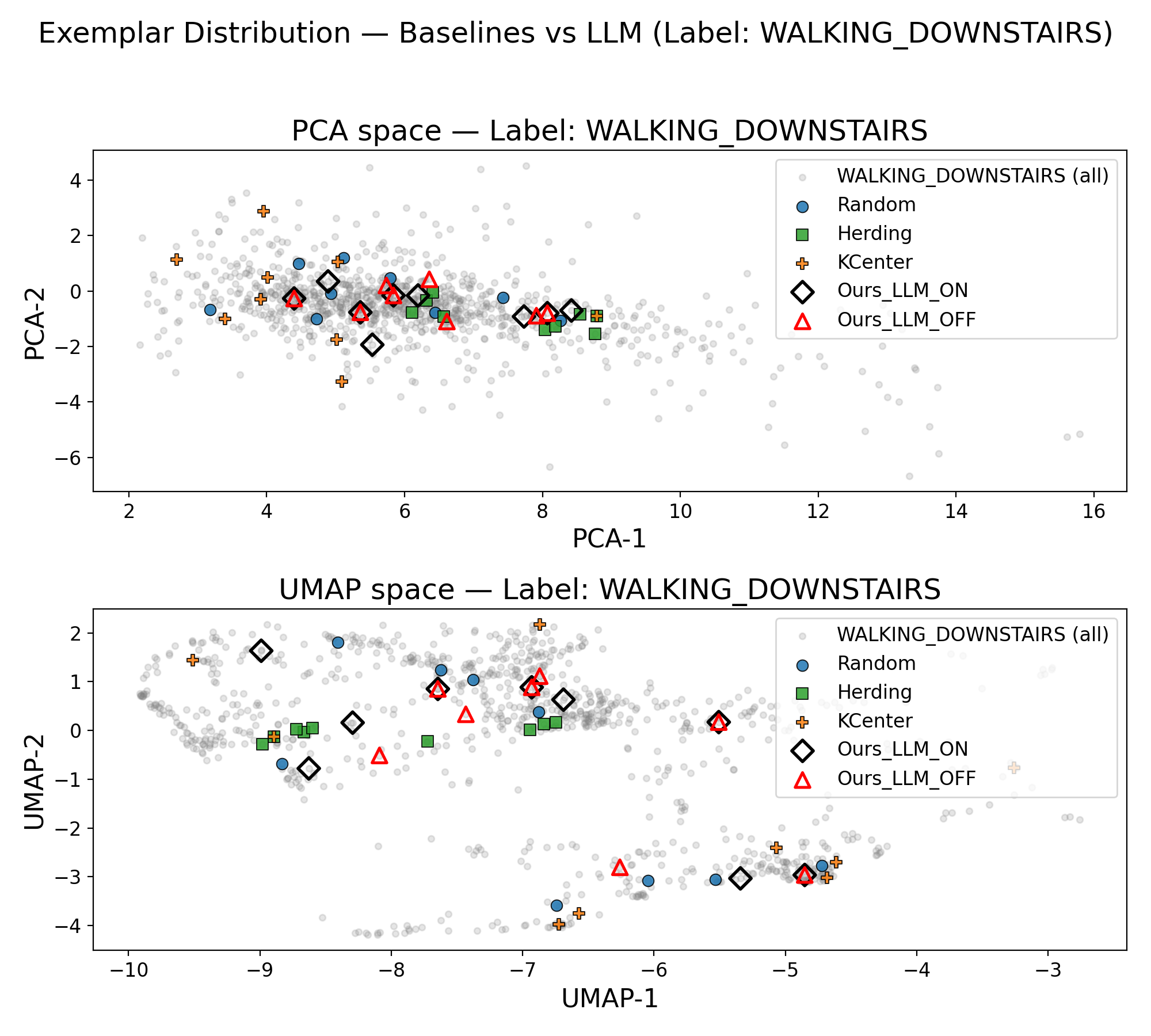} &
\includegraphics[width=0.5\linewidth, keepaspectratio]{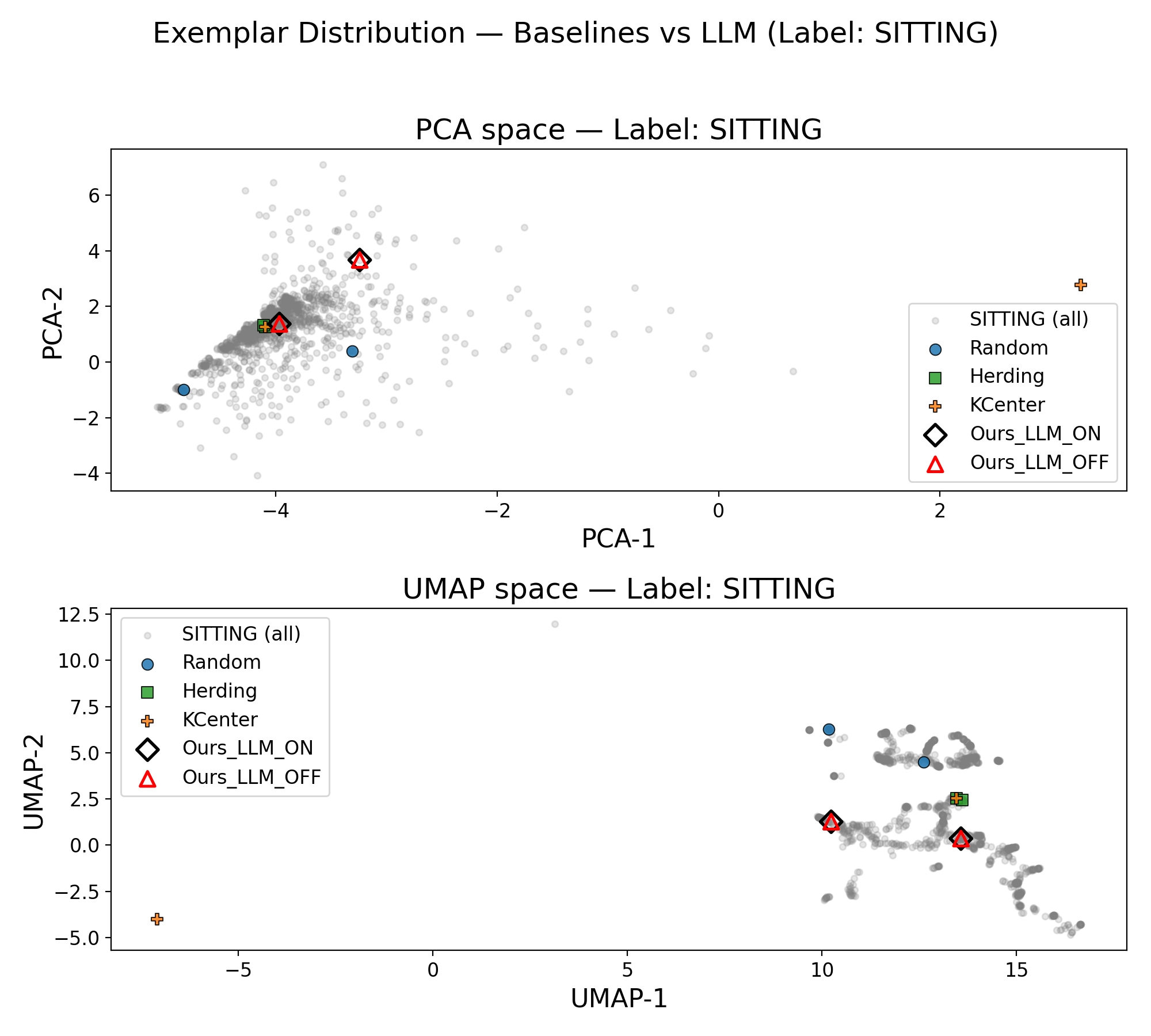}\\[-1pt]
(c) Walking Downstairs & (d) Sitting\\[4pt]
\includegraphics[width=0.5\linewidth, keepaspectratio]{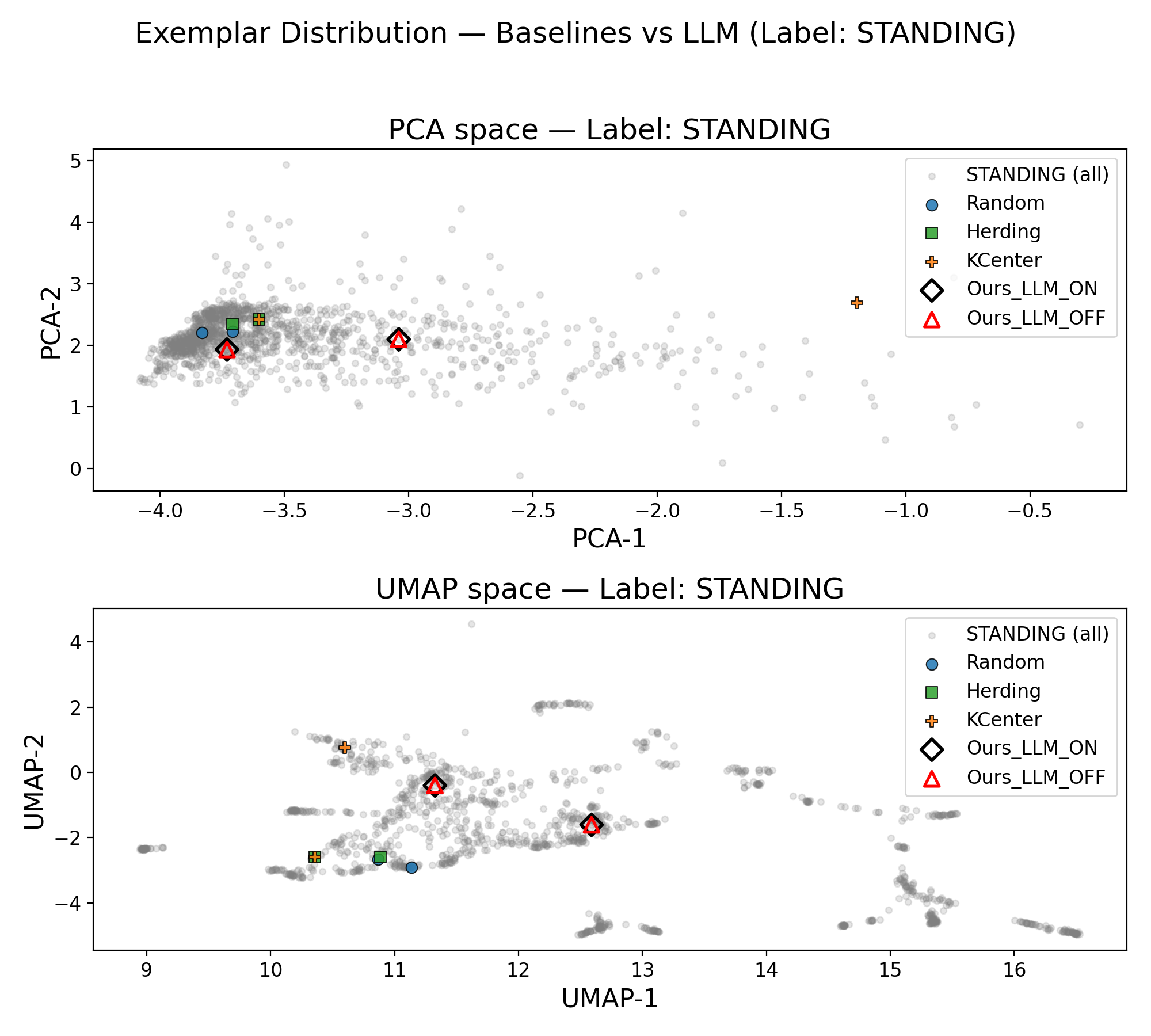} &
\includegraphics[width=0.5\linewidth, keepaspectratio]{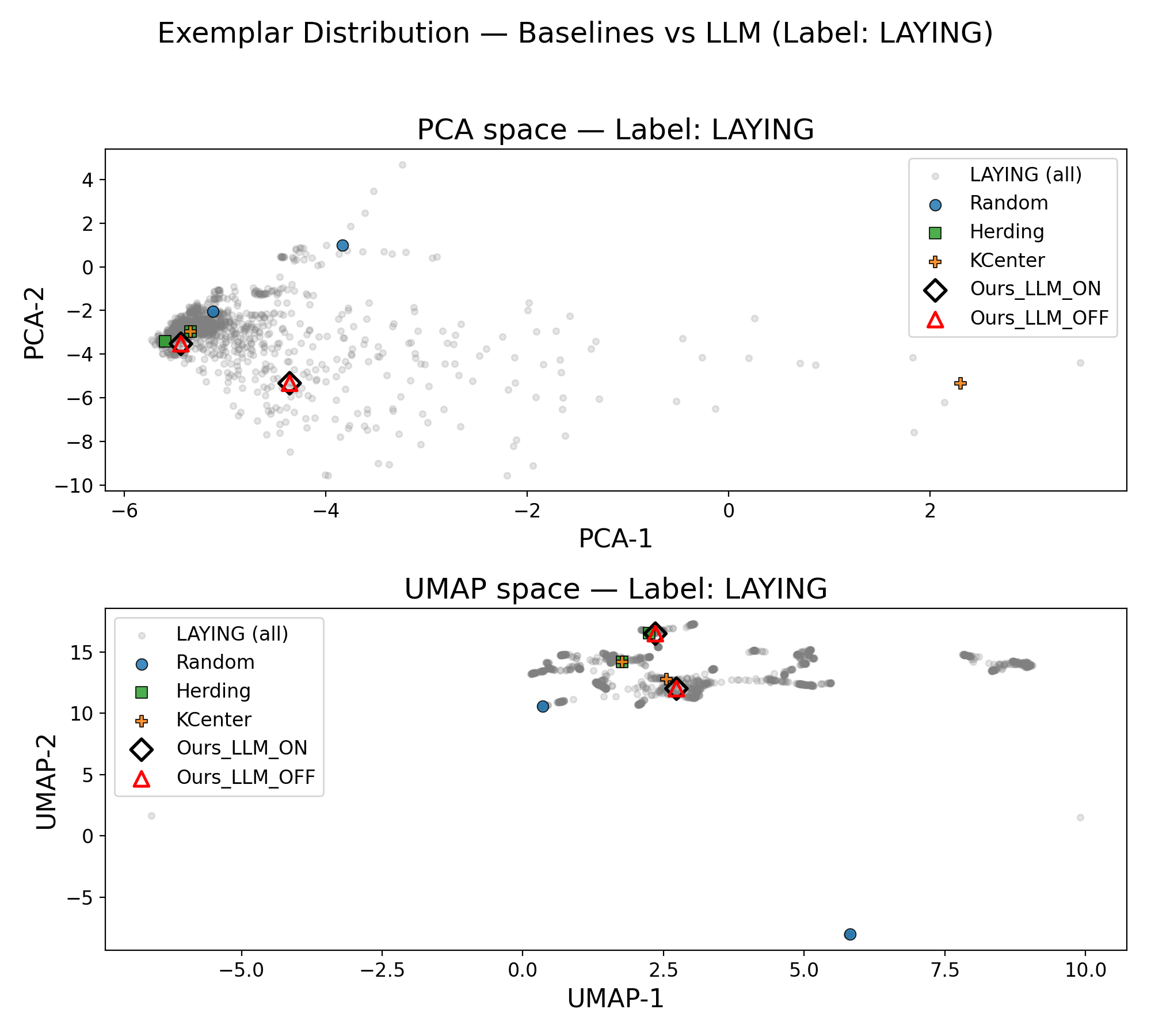} \\[-1pt]
(e) Standing & (f) Laying
\end{tabular}
\caption{Per-class UMAP visualizations showing exemplar distributions for each activity.}
\label{fig:umap-per-class-llm}
\end{figure}

Figure~\ref{fig:umap-per-class-llm} presents UMAP visualizations for each activity label. Gray points represent all training samples for that label, while colored markers denote exemplars from each selection method.
Main findings:
\begin{itemize}
    \item For dynamic activities (WALKING, WALKING\_UPSTAIRS, WALKING\_DOWNSTAIRS), LLM-ON captures subclusters associated with different gait patterns.
    \item For static activities (SITTING, STANDING, LAYING), LLM-ON places exemplars near stable low-variance regions, consistent with the semantic priors.
    \item Across classes, LLM-ON avoids redundant or ambiguous regions more effectively than LLM-OFF and baselines.
\end{itemize}
The UMAP visualizations provide qualitative support for the role of LLM priors in exemplar selection. Compared to baseline methods and to the LLM-OFF variant, the LLM-guided approach consistently selects exemplars that align with semantically meaningful subregions of the data manifold. These patterns complement the quantitative improvements observed in the main experimental results.
\FloatBarrier
\FloatBarrier

\section{Appendix: Time and computational cost}
\label{time}
\begin{figure}[!ht]
\centering
\includegraphics[width=0.95\linewidth]{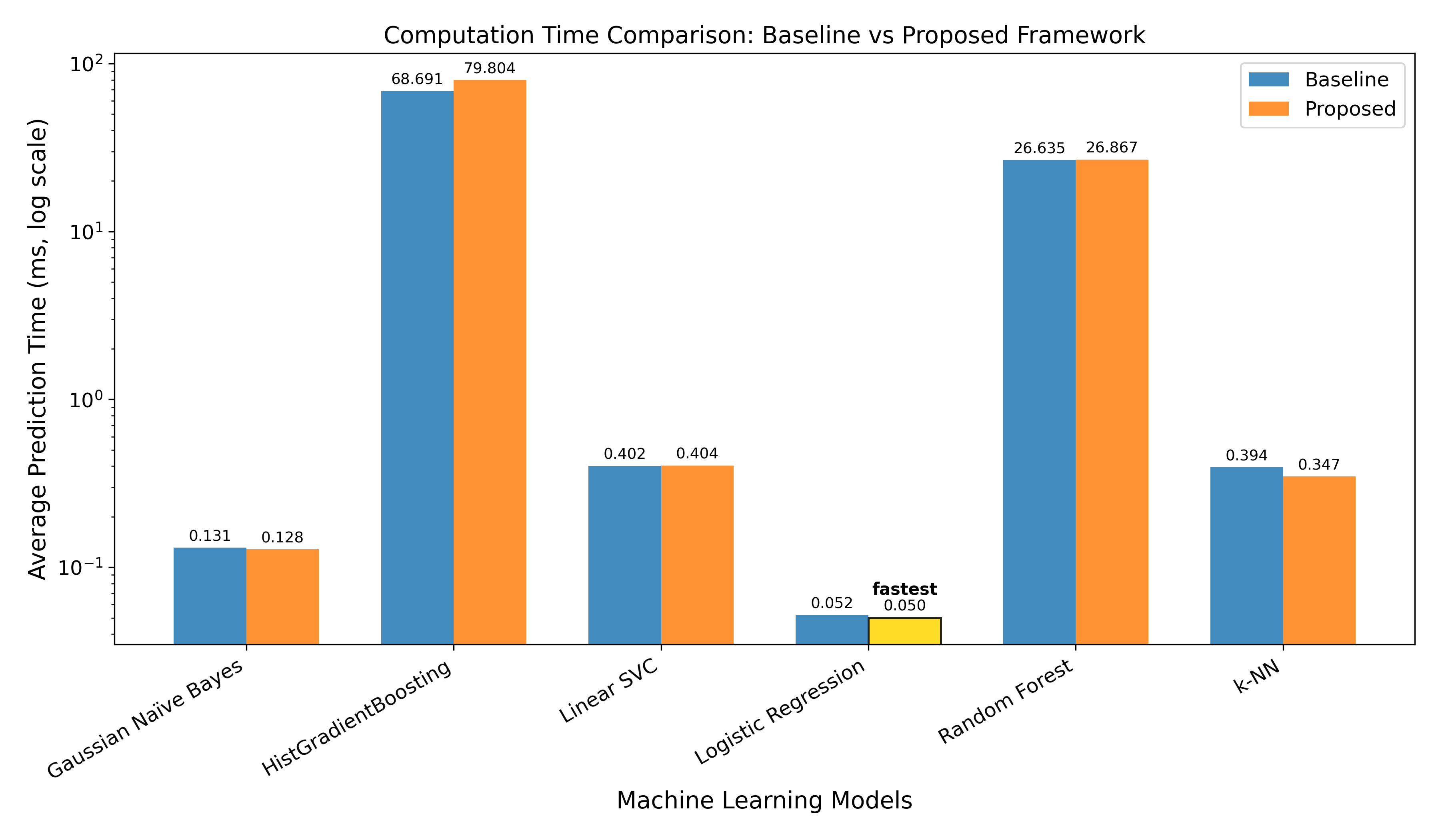}
\caption{
Computation time comparison between baseline exemplar selection and the proposed LLM-guided framework across multiple machine-learning models. The y-axis uses a logarithmic scale to visualize differences among fast and slow models. Logistic Regression achieves the fastest prediction time, highlighted in gold. Overall, the proposed method maintains inference speed comparable to the baseline, indicating that incorporating LLM-derived exemplar selection does not introduce additional runtime overhead.
}
\label{fig:appendix_time}
\end{figure}

To assess the computational efficiency of the suggested framework, we compared the average prediction time per test sample between the baseline exemplar-selection methods and our LLM-Guided approach. Figure~\ref{fig:appendix_time} presents the runtime of several classical machine-learning classifiers, including Logistic Regression, Random Forest, HistGradientBoosting, Linear SVC, k-NN, and Gaussian Naïve Bayes. A logarithmic scale is used on the y-axis to visualize differences between fast and slow models clearly.

In general, the findings indicate that the suggested framework has minimal computational overhead compared to the baseline. For most models, the prediction time remains nearly identical across the two configurations, indicating that the LLM-guided exemplar selection does not introduce additional latency during inference. The only differences observed are minor fluctuations in the millisecond range, which are negligible for real-time or embedded wearable-sensor applications.

Among all evaluated models, Logistic Regression achieves the fastest prediction time, highlighted in gold in the figure. This model requires only 0.050 ms/sample in the proposed framework, slightly faster than the baseline (0.052 ms/sample). Gaussian Naïve Bayes and k-NN also demonstrate low latency, while more complex models such as HistGradientBoosting and Random Forest exhibit higher runtimes due to their inherent algorithmic complexity.

These findings support the claim that the proposed exemplar-selection process improves classification performance without sacrificing inference speed, making it suitable for deployment in lightweight or resource-constrained sensor-based HAR systems.

\end{document}